\documentclass{article}

\usepackage[final]{neurips_2024}

\usepackage{natbib}
\setcitestyle{numbers,square}

\usepackage{graphicx}
\usepackage{subfigure}

\usepackage{amsmath}
\usepackage{amssymb}
\usepackage{mathtools}
\usepackage{amsthm}

\usepackage{multirow}
\usepackage{import}
\usepackage{pifont}
\usepackage{textcomp}
\usepackage{color, colortbl}
\definecolor{greyC}{RGB}{180,180,180}
\definecolor{greyL}{RGB}{235,235,235}
\usepackage{footmisc}
\usepackage[noend]{algpseudocode}
\usepackage{bm}
\usepackage[flushleft]{threeparttable}
\usepackage[nospace]{cite}
\usepackage[]{hyperref} 
\usepackage{makecell}

\makeatletter
\newcommand*{\addFileDependency}[1]{
  \typeout{(#1)}
  \@addtofilelist{#1}
  \IfFileExists{#1}{}{\typeout{No file #1.}}
}
\makeatother

\usepackage[utf8]{inputenc} 
\usepackage[T1]{fontenc}    
\usepackage{hyperref}       
\usepackage{url}            
\usepackage{booktabs}       
\usepackage{amsfonts}       
\usepackage{nicefrac}       
\usepackage{microtype}      
\usepackage{xcolor}         
\usepackage{adjustbox}
\usepackage{subcaption}
\usepackage{enumitem}

\usepackage{hyperref}
\usepackage{algorithm}
\usepackage{algcompatible}
\usepackage{wrapfig}
\theoremstyle{plain}
\newtheorem{theorem}{Theorem}[section]
\newtheorem{proposition}[theorem]{Proposition}
\newtheorem{lemma}[theorem]{Lemma}
\newtheorem{corollary}[theorem]{Corollary}
\theoremstyle{definition}
\newtheorem{definition}{Definition}[section]
\newtheorem{assumption}{Assumption}[section]
\theoremstyle{remark}
\newtheorem{remark}{Remark}[section]

\newcommand{\unishrinkA}{\vspace{-0.0cm}}
\newcommand{\unishrinkB}{\vspace{-0.0cm}}

\newcommand{\cmark}{\ding{51}}%
\newcommand{\xmark}{\ding{55}}%

\usepackage{amsmath,amsfonts,bm}

\def\eqref#1{equation~\ref{#1}}

\def\1{\bm{1}}

\usepackage{mathtools}
\usepackage{amssymb}
\usepackage{latexsym}
\usepackage{dsfont}
\usepackage{mathrsfs}
\usepackage{amssymb}
\usepackage{amsfonts}
\usepackage{bm}
\usepackage{xspace}
\usepackage{amsthm}
\usepackage{multirow}

\newcommand{\conv}{\mathrm{conv}}

\newcommand{\Dcal}{\mathcal{D}}

\newcommand{\Mcal}{\mathcal{M}}

\newcommand{\Scal}{\mathcal{S}}

\newcommand{\Ycal}{\mathcal{Y}}

\ifx\BlackBox\undefined
\newcommand{\BlackBox}{\rule{1.5ex}{1.5ex}}  
\fi

\ifx\QED\undefined
\def\QED{~\rule[-1pt]{5pt}{5pt}\par\medskip}
\fi

\ifx\proof\undefined
\newenvironment{proof}{\par\noindent{\em Proof:\ }}{\hfill\BlackBox\\[.0mm]}
\fi

\ifx\theorem\undefined
\newtheorem{theorem}{Theorem}[section]
\fi

\ifx\example\undefined
\newtheorem{example}{Example}[section]
\fi

\ifx\property\undefined
\newtheorem{property}{Property}
\fi

\ifx\lemma\undefined
\newtheorem{lemma}{Lemma}[section]
\fi

\ifx\proposition\undefined
\newtheorem{proposition}{Proposition}[section]
\fi

\ifx\remark\undefined
\newtheorem{remark}{Remark}
\fi

\ifx\corollary\undefined
\newtheorem{corollary}{Corollary}[section]
\fi

\ifx\definition\undefined
\newtheorem{definition}{Definition}[section]
\fi

\ifx\conjecture\undefined
\newtheorem{conjecture}{Conjecture}[section]
\fi

\ifx\axiom\undefined
\newtheorem{axiom}[theorem]{Axiom}
\fi

\ifx\claim\undefined
\newtheorem{claim}[theorem]{Claim}
\fi

\ifx\assumption\undefined
\newtheorem{assumption}{Assumption}
\fi

\ifx\problem\undefined
\newtheorem{problem}{Problem}[section]
\fi

\ifx\fact\undefined
\newtheorem{fact}{Fact}
\fi

\newcommand{\sbr}[1]{\left[#1\right]}
\newcommand{\cbr}[1]{\left\{#1\right\}}

\newcommand{\benr}{\begin{eqnarray}}
\newcommand{\eenr}{\end{eqnarray}}
\newcommand{\benrr}{\begin{eqnarray*}}
\newcommand{\eenrr}{\end{eqnarray*}}
\newcommand{\ben}{\begin{equation}}
\newcommand{\een}{\end{equation}}
\newcommand{\benn}{\begin{equation*}}
\newcommand{\eenn}{\end{equation*}}

\def\size{{S}}
\def\Fens{{F_\text{ens}}}
\def\Dtest{{\Dcal_{\text{test}}}}

\def\ttest{\text{test}}
\def\tloc{\text{loc}}
\def\tfus{\text{fus}}
\def\tback{\text{back}}
\def\conv{conv1$\times$1}

\newcommand{\Floc}[1]{F^\text{loc}_#1}
\newcommand{\Ffus}[1]{F_{\text{fus}}^{#1}}
\newcommand{\Hfus}[1]{H_\text{fus}^{#1}}

\newcommand{\SPU}[1]{R_{#1}^{\text{spu}}}
\newcommand{\INV}[1]{R_{#1}^{\text{inv}}}
\newcommand{\dSPUloc}[1]{d_{\text{loc},#1}^{\text{spu}}}
\newcommand{\dINVloc}[1]{d_{\text{loc},#1}^{\text{inv}}}

\newcommand{\Ouralgo}{\texttt{FuseFL}}

\title{FuseFL: One-Shot Federated Learning through the Lens of Causality with Progressive Model Fusion}

\author{\hspace{-1mm}
Zhenheng Tang\textsuperscript{$\dag$}\thanks{This work is partially done during the visiting in The Hong Kong University of Science and Technology (Guangzhou).} \quad Yonggang Zhang\textsuperscript{$\dag$} \quad 
Peijie Dong\textsuperscript{$\sharp$} \quad 
Yiu-ming Cheung\textsuperscript{$\dag$} \\
\bf Amelie Chi Zhou\textsuperscript{$\dag$} \quad 
Bo Han\textsuperscript{$\dag$} \quad 
Xiaowen Chu\textsuperscript{$\sharp, \S$} \\
\textsuperscript{$\dag$} Department of Computer Science, Hong Kong Baptist University  \\
\textsuperscript{$\sharp$} DSA Thrust, The Hong Kong University of Science and Technology (Guangzhou)  \\
\texttt{\{zhtang, ygzhang, ymc, amelieczhou, bhanml\}@comp.hkbu.edu.hk} \\
\texttt{\{pdong212, xwchu\}@connect.hkust-gz.edu.cn}
}

\begin{document}

\maketitle
\begingroup\renewcommand\thefootnote{$^{\S}$}
\footnotetext{Correspondence to Xiaowen Chu (xwchu@hkust-gz.edu.cn).}
\endgroup


\begin{abstract}

One-shot Federated Learning (OFL) significantly reduces communication costs in FL by aggregating trained models only once. However, the performance of advanced OFL methods is far behind the normal FL. In this work, we provide a causal view to find that this performance drop of OFL methods comes from the isolation problem, which means that locally isolatedly trained models in OFL may easily fit to spurious correlations due to data heterogeneity. From the causal perspective, we observe that the spurious fitting can be alleviated by augmenting intermediate features from other clients. Built upon our observation, we propose a novel learning approach to endow OFL with superb performance and low communication and storage costs, termed as \Ouralgo{}. Specifically, \Ouralgo{} decomposes neural networks into several blocks and progressively trains and fuses each block following a bottom-up manner for feature augmentation, introducing no additional communication costs. Comprehensive experiments demonstrate that \Ouralgo{} outperforms existing OFL and ensemble FL by a significant margin. We conduct comprehensive experiments to show that \Ouralgo{} supports high scalability of clients, heterogeneous model training, and low memory costs. Our work is the first attempt using causality to analyze and alleviate data heterogeneity of OFL\footnote{The code is publicly available: \url{https://github.com/wizard1203/FuseFL}}.
\end{abstract}

\unishrinkA{}
\section{Introduction}
Federated learning (FL)~\citep{mcmahan2017communication,kairouz2021advances} has become a popular paradigm that enables collaborative model training without sharing private datasets from clients. Two typical characteristics of FL limit its performance: (1) FL normally has non-IID (Independently and Identically Distributed) data, also called \textit{data heterogeneity}~\citep{kairouz2021advances}, which causes unstable slow convergence~\citep{karimireddy2019scaffold,woodworth2020minibatch,pmlr-v202-sun23h,tang2024fedimpro} and poor model performance~\citep{yuan2022what,DFL,VHL,zhangrobust,yang2024fedfed}; (2) The extremely low bandwidth, e.g. $1\sim 10$ MB/s of FL in Internet environments~\citep{kairouz2021advances,tang2023fusionai,tang2024fusionllmdecentralizedllmtraining,BCRS-OPWA,GossipFL,tang2024fusionllmdecentralizedllmtraining}, leads to high communication time of a large neural network. For example, communicating once ResNet-50~\citep{resnet} with 25.56M parameters (102.24MB) or GPT-3~\citep{GPT3} with 175B parameters (700GB) will consume around 102.24 seconds or 194 hours, respectively. Current FL methods alleviate this problem by skipping the gradient synchronization of traditional distributed training to save communication costs~\citep{mcmahan2017communication,kairouz2021advances}. But the required hundreds or thousands of communication rounds still make the communication time unacceptable.

To reduce the communication costs at extreme, one-shot FL (OFL)~\citep{zhang2022dense,guha2019one,li2020practical,zhou2020distilled,dennis2021heterogeneity,dai2024enhancing} only communicates the local trained model once. Thus, the communication cost is the model size $\size$ for each client, less than FedAvg-style algorithms for hundreds or thousands of times. However, averaging for only once cannot guarantee the convergence of FedAvg. Thus, the direct idea is to aggregate client models on the server and conduct inference as ensemble learning does. Some advanced works also consider better model averaging~\citep{jhunjhunwala2023towards,qu2022generalized,liu2023bayesian,al2020federated}, neurons matching~\citep{ainsworth2022git, wang2020Federated}, selective ensemble learning~\citep{diao2023towards,heinbaugh2023datafree,wang2023datafree}, model distillation~\citep{li2020practical,zhou2020distilled,dennis2021heterogeneity,dai2024enhancing}. These methods may be impractical due to the requirements of additional datasets with privacy concerns, and the extra large storage or computation costs. Most importantly, there still exists a large performance gap between OFL and the normal FL or the ensemble learning. This motivates the following question:

\begin{quote}
\emph{How to improve FL performance under \textbf{extremely low communication costs} with almost no extra computational and storage costs?}
\end{quote}

In this work, we provide a \textit{causal view}~\citep{pearl2009causalityResnet,peters2017elements,ahuja2021invariance,scholkopf2021toward} to analyze the performance drop of OFL. We firstly construct a causal graph to model the data generation process in FL, where the spurious features build up the data heterogeneity between clients, and invariant features of the same class remain constant in each client (domain)~\citep{ahuja2021invariance,scholkopf2021toward,chen2023understanding,ye2021towards,zhang2021adversarial}. Then, we show the performance drop comes from the \textit{isolation problem}, which means that locally isolatedly trained models in OFL may easily fit to spurious correlations like adversarial shortcuts~\citep{adv_attack,geirhos2020shortcut,hendrycks2021natural}, instead of learning invariant features~\citep{ahuja2021invariance,scholkopf2021toward}, causing a performance drop of OFL on the test dataset. Consider a real-world example, Alice takes photos of birds in the forests, while Bob near the sea. Now, the isolated models will mistakenly identify birds according to the forests or the sea~\citep{hendrycks2021natural,geirhos2020shortcut}. Based on the causal graph, we intuitively and empirically show that such spurious fitting can be alleviated by augmenting intermediate features from other clients (Section~\ref{sec:analysis}).  



Built upon this observation, we propose a simple yet effective learning approach to realizing OFL with superb performance and extremely low communication and storage costs, termed as \Ouralgo{}, which builds up the global model through bottom-up training and fusion to improve OFL performance (Section~\ref{sec:method}). Specifically, we split the whole model into multiple blocks~\footnote{Note that this method is general to any neural network, and the fusion is different from averaging.} (The ``block'' means a single or some continuous layers in a DNN.). For each block, clients first train and share their local blocks with others; then, these trained local blocks are assembled together, and the features outputted from these blocks are fused together and fed into the next local blocks. This process is repeated until the whole model is trained. Through this bottom-up training-and-fusion method, local models can learn better feature extractors that learn more invariant features from other clients to avoid the isolation problem. To avoid the large storage cost, given the number of clients $M$, we assign each local client with a small model with reduced hidden dimension with ratio $\sqrt{M}$, to ensure the final learned global model has the same size $\size$ as the original model.

Our main contributions can be summarized as follows:
\unishrinkA{}
\begin{itemize}[leftmargin=*]
    \item \noindent We provide a causal view to understand the gap between multi-round FL and OFL, showing that augmenting intermediate features from other clients contributes helps improve OFL. As far as we know, this is the first work using causality to analyze the data heterogeneity of OFL.
    \item \noindent To leverage causality to improve OFL, we design \Ouralgo{}, which decomposes models into several modules and transmits one module for feature augmentation at each communication round.
    \item \noindent We conduct comprehensive experiments to show how \Ouralgo{} significantly promotes the performance of OFL without \emph{no} additional communication and computation cost.
\end{itemize}





\unishrinkA{}
\section{Preliminary}\label{sec:back}
\unishrinkA{}
\subsection{Federated Learning}\label{sec:FL}
\unishrinkA{}



In FL, a set of clients $\Mcal = \left\{m| m\in 1,2,...,M  \right\}$ have their own dataset $\Dcal_m$. Given $C$ classes indexed by $[C]$, a sample in $\Dcal_m$ is denoted by $(x,y)\in \mathcal{X}\times[C]$, where $x$ is the input in the space $\mathcal{X}$ and $y$ is its corresponding label. These clients cooperatively learn a model $F(\theta, x) : \mathcal{X} \to \mathbb{R}^C$ that is parameterized as $\theta \in \mathbb{R}^d $. Formally, the global optimization problem can be formulated as~\citep{mcmahan2017communication}:
\begin{equation}\nonumber\label{eq:F}
    \min_{\theta} L(\theta) \triangleq \sum_{m=1}^{M} p_m L_m(\theta)    = \sum_{m=1}^{M} p_m \mathbb{E}_{(x,y)\sim \Dcal_m} \ell(F;x,y),
\end{equation}


where the local objective function of $m$th-client $\ell(F;x,y )\triangleq CE(\hat{y},y)$ with $\hat{y}\triangleq F(\theta;x)$, $CE$ denotes the cross-entropy loss, $p_m > 0 $ and $ \sum_{m=1}^{M} p_m =1  $. Usually, $p_m\triangleq {n_m}/{N}$, where $n_m$ denotes the number of samples on client $m$ ($n_m=|\Dcal_m|$) and $N=\sum_{m=1}^{M} n_m$.

The classic FL algorithm is FedAvg~\citep{mcmahan2017communication}. In each communication round $t$, the central server randomly samples a part of clients $\Scal^{t} \subseteq \Mcal$ and broadcasts the model $\theta^{t}$ to all selected clients, and then each $m$-th client performs multiple local updates. After local training, all selected clients send the optimized $\theta_{m, E}^{t}$ to the server, and the server aggregates and averages local models to obtain a global model. Such a multi-round communication introduces large communication costs~\citep{kairouz2021advances}.



\unishrinkA{}
\subsection{Ensembled FL}\label{sec:ensemble}
\unishrinkA{}
The FedAvg requires multiple communication rounds $T$ for convergence~\citep{woodworth2020minibatch,li2019convergence}, which might be extremely large~\citep{kairouz2021advances}. Given the model size $\size$, FedAvg-style FL methods introduce communication costs as $T\times\size$. As shown in Table~\ref{tab:demystify}, the current lowest communication cost of FL is reduced as $\size$ in OFL, making FL possibly deployable in low communication bandwidth scenarios~\citep{tang2023fusionai,tang2024fusionllmdecentralizedllmtraining}. Thus, we analyze what causes the performance drop of OFL and how to improve it. As the performance of average-based and model distillation OFL methods is upper bounded by ensemble learning~\citep{zhang2022dense,guha2019one,li2020practical,zhou2020distilled,dai2024enhancing}, we mainly focus on analyzing ensemble learning and differentiating \Ouralgo{} from it. The output of the ensemble learning can be formalized as: $\Fens(x) \triangleq \frac{1}{M} \sum_{m\in \Mcal} \Floc{m}(\theta_m;x)$, in which the local model $\Floc{m}$ parameterized with $\theta_m$ is isolatedly trained with minimizing empirical risk minimization (ERM) objective~\citep{OODgeneralization,ahuja2021invariance,chen2023understanding} function $\ell(F(\theta, x) ,y), (x,y) \sim \Dcal_m$ by SGD.


\unishrinkA{}
\section{Federated Learning: A Causal View}\label{sec:analysis}
\unishrinkA{}

\begin{wrapfigure}{R}{0.5\textwidth}
\begin{minipage}{0.5\textwidth}
\begin{figure}[H]
\unishrinkB{}
    \setlength{\belowdisplayskip}{2pt}
    \setlength{\abovedisplayskip}{-5pt}
   \centering
    {\includegraphics[width=1.0\linewidth]{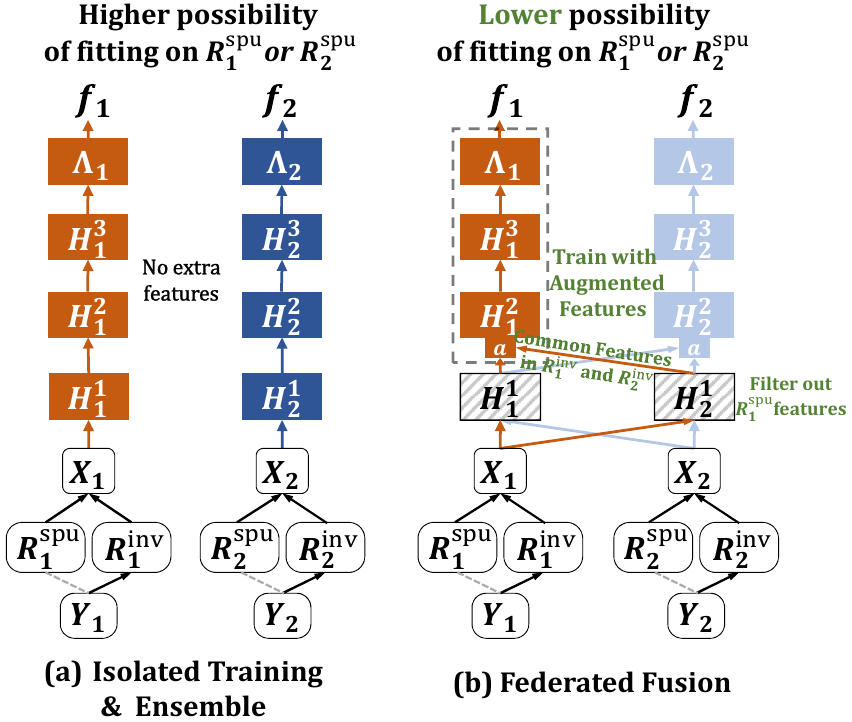}}
    \unishrinkB{}
    \caption{Structure Equation Model~\citep{pearl2009causalityResnet} of FL.}
    \label{fig:Motivation}
\unishrinkB{}
\end{figure}
\end{minipage}
\end{wrapfigure}


\subsection{The Sequential Structure of Neural Networks}\label{sec:NN}
\unishrinkA{}
A neural network can be decomposed into the sequential module-wise structure as shown in Figure~\ref{fig:Motivation}. Formally, it can be defined as:
\begin{equation}\label{eq:seqNN}
    F = \Lambda \circ H^K \circ H^{K-1} \circ \cdots \circ H^{1},   \text{for} \ 1 \leq k \leq K,
\end{equation}
where $\Lambda$ is the final classifier, and $H_k$ is the module that may consist of single or multiple blocks. The $\Lambda$ and each $H_k$ are parameterized by $\theta^{\Lambda} \in \mathbb{R}^{d_{\Lambda}}$ and $\theta^{k} \in \mathbb{R}^{d_k}$. The $H^{i} \circ H^{j} (\cdot)$ means $H^{i}(H^{j} (\cdot))$.
Thus, the parameter $\theta$ of $F$ are concatenated by the $\theta^{\Lambda}$ and $\left\{ \theta^{k}|k\in 1,2,\cdots K \right\}$, and $d_{\Lambda} + \sum_{k=1}^K d_k = d$. 

As Figure~\ref{fig:Motivation} illustrates, each module $H^{k}$ receives the output of module $H^{k-1}$, and the final classifier receives the output of the final hidden module $H^{K}$ and makes predictions $\hat{y}=f(x)$ on the input $x$. We call the output from each module, $h^{k} = H^{k}(h^{k-1})$ and $h^{1} = H^{1}(x)$, as the feature for simplicity. 

\subsection{Structure Equation Model of FL}\label{sec:SEM}
Inspired from the analysis of out-of-distribution (OOD) generalization~\citep{ahuja2021invariance} through the lens of mutual information~\citep{shwartz2017opening,achille2018emergence} and structure equation model (SEM) in causality~\citep{peters2017elements,ahuja2021invariance,scholkopf2021toward,zhang2021adversarial}, we define the data generation SEM of FL as shown in Figure~\ref{fig:Motivation}. For local training dataset $\Dcal_m$ at client $m$, the SEM is $Y_m \to \INV{m} \to X_m \gets \SPU{m}$, where $\INV{m}$ and $\SPU{m}$ are invariant and spurious features, $Y_m$ and $X_m$ are label and input data respectively. Here, the dataset $\Dcal_m$ is a subset of the whole dataset $\Dcal$. The $\SPU{m}$ is actually the nuisance at a global level (respect to $Y$), being independent of $Y$, but dependent on $X_m$.


\textbf{Non-IID data and causality.} 
For a groundtruth label $Y$, its corresponding invariant features $R_{m}^{\text{inv}}$ do not change across clients~\citep{ahuja2021invariance,scholkopf2021toward,chen2023understanding,ye2021towards}. However, the spurious features $R_{m}^{\text{spu}}$ are other factors that occasionally exist in data and do not have a relationship to $Y$, which means that the heterogeneous features of data (non-iid) with the same class come from the spurious features $R_{m}^{\text{spu}}$ (concept shift~\citep{kairouz2021advances}).
For example, in photos of birds in the forests or the sea, pixels of birds are $R_{m}^{\text{inv}}$ while the forests and the sea are  $R_{m}^{\text{spu}}$. Considering the test dataset includes all client data distribution and even OOD data, the $Y_{\ttest}$ is largely dependent on $\INV{\ttest}$: $P(Y_\ttest|X_{\ttest}, \INV{\ttest}) \gg P(Y_\ttest|X_{\ttest}, \SPU{\ttest}) $\footnote{Without loss of generality, one can also model the SEM of test data as $Y_\ttest \to \INV{\ttest} \to X_\ttest$ for simplicity.}.

\textbf{Non-IID scenarios.} This SEM model considers the label shift ($p_i(y)\neq p_j(y)$) and concept shift ($p_i(x|y)\neq p_j(x|y)$) scenarios~\citep{kairouz2021advances,li2021federatedstudy}, or both of them appear simultaneously. When the support\footnote{Support is the region where the probability density for continuous random variables (probability mass function for discrete random variables) is positive~\citep{ahuja2021invariance}.} $\Ycal_m$ of $Y_m$ is different or partly overlapped between clients $m=1,...,M$, this would be the severe non-IID scenario~\citep{yu2020federated}. And it is obvious that spurious features $\SPU{m}$ relate to the concept shift.

\textbf{Spurious fitting.} By conducting isolated local training on local dataset $\Dcal_m$ at client $m$, the model $\Floc{m}$ is prone to learn to predict $Y_m$ based on spurious features $\SPU{m}$, i.e. low distance $\dSPUloc{m} = d(P(Y_m|X_m, \SPU{m}), P(\Floc{m}|X_m, \SPU{m}))$ but high distance $d(P(Y_m|X_m, \INV{m}), P(\Floc{m}|X_m, \INV{m}))$, in which the distance $d$ could be $CE$ loss or $KL$ divergence. The reason for the spurious fitting by isolated training is that the invariant features $\INV{i\neq m}$ from other clients are not observed by client $m$, while the $\SPU{m}$ frequently appears in the local dataset $\Dcal_m$ like the adversarial attacks or shortcuts~\citep{adv_attack,geirhos2020shortcut,hendrycks2021natural}. This guarantees low error on the training dataset $\Dcal_m$, because it has much less data than $\Dcal_{\ttest}$ and $\Dcal_{1,...,M}$, thus introducing high probability $P(Y_m|X_m, \SPU{m})$. However, on test dataset $\Dtest$, the low $\dSPUloc{m}$ of model $\Floc{m}$ but high $\dINVloc{m}$ of model $\Floc{m}$ leads to high test error. Different from isolated training, FedAvg alleviates this problem by multiple times of averaging models to find those common features, including more $\INV{1,...,M}$ and removing $\SPU{1,...,M}$.


\textbf{Feature augmentation.} Through the above analysis, the key to improve OFL performance is to endow OFL with the ability of training to see invariant features across all clients. It has been found that training on noised datasets with SGD to optimize ERM can still result in some feature representations consisting of both spurious and invariant features; exploiting the invariant features is the key to helping improve OOD performance~\citep{zhang2021can,ahuja2021invariance,arjovsky2019invariant}. In light of this, we introduce augmenting features by fusing client models block by block. Concisely speaking, in \Ouralgo{}, each local model can conduct local training with the view from other clients $H_{i\neq m}(X_m)$, which helps filter out $\SPU{m}$ but retain $\INV{m}$, as other clients cannot see $\SPU{m}$ in their dataset $\Dcal_{i \neq m}$. This method can be seen as a kind of invariant feature augmentation~\citep{chen2023understanding}. The details of \Ouralgo{} are shown in Section~\ref{sec:method}.

\unishrinkA{}
\subsection{Mutual Information}\label{sec:MI}
\unishrinkA{}
The goal of FL is to obtain a model that performs well on all client datasets~\citep{kairouz2021advances}. Thus, here we consider the random variable $X,Y$ sampled from the global dataset $\Dcal$. In this section, we also write $H^k$ as the features that output from $H^k(H^{k-1}\dots( H^1(X)))$ for simplicity.

Given the probabilistic graph model $(\SPU{}, Y) \to X \to H^1 \to \cdots \to H^k \to F(X)$ (Eq.~\ref{eq:seqNN}), where $\INV{}$ are ignored for simplicity, the MI between $Y$ and subsequent transformations $H^k$ on $X$ satisfies a decreasing trend: $I(X; Y)\geq I(H^1; Y)\geq \cdots \geq I(H^K; Y)$; the MI between $X$ and subsequent transformations on $X$ satisfies a decreasing trend: $Entropy(X) \geq I(H^1; X)\geq \cdots \geq I(H^K; X)$~\citep{shwartz2017opening}. If $I(H^K; \SPU{})=0$ and $H^K$ can predict labels, $H^K$ is called invariant features so that the final classifier will not overfit to spurious correlations between $\SPU{}$ and $Y$. The previous works~\citep{shwartz2017opening} show that achieving the following minimal sufficient statistic provides good generalization:
\begin{align}   
\text{\textbf{Sufficient statistic}:}  & \ I(X; Y) = I(H(X); Y), \label{eq:IB-Y} \\ 
\text{\textbf{Minimal statistic}:}  & \ H(X) = \arg \min_{\Tilde{H}(X)} I(\Tilde{H}(X); X). \label{eq:IB-X}
\end{align}

\begin{lemma}[Invariance and minimality~\citep{achille2018emergence}]\label{lemma:invariance}
Given spurious feature $\SPU{}$ for the label $Y$, and probabilistic graph model $(\SPU{}, Y) \to X \to H(X)$, then,
\begin{equation}\nonumber
    I(H(X); \SPU{}) \leq I(H(X);X) - I(X;Y).
\end{equation}
There is a nuisance $\SPU{}$ such that equality holds up to a residual $\epsilon$: 
\begin{equation}\nonumber
    I(H(X); \SPU{}) = I(H(X);X) - I(X;Y) - \epsilon,
\end{equation}
where $\epsilon \triangleq I(H(X);Y|\SPU{}) - I(X;Y)$. The sufficient statistic $H(X)$ (satisfying Eq.~\ref{eq:IB-X}) is invariant to $\SPU{}$ if and only if it is minimal (satisfying Eq.~\ref{eq:IB-Y}).
\end{lemma}
\begin{remark}
Based on Lemma~\ref{lemma:invariance}, we can study how $I(H(X); X)$ and $I((H);Y)$ changes to study to what degree the $H(X)$ contains spurious features.
\end{remark}



\begin{figure*}[htb!]
    \subfigbottomskip=-1pt
    \subfigcapskip=1pt
  \centering
     \subfigure[Estimated MI $I(H^k;X)$.]{\includegraphics[width=0.3\textwidth]{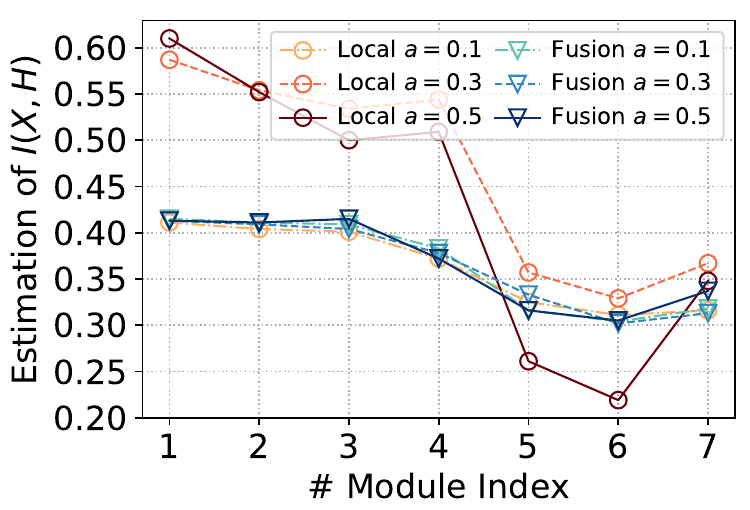}}
     \subfigure[Estimated MI $I(H^k;Y)$.]{\includegraphics[width=0.3\textwidth]{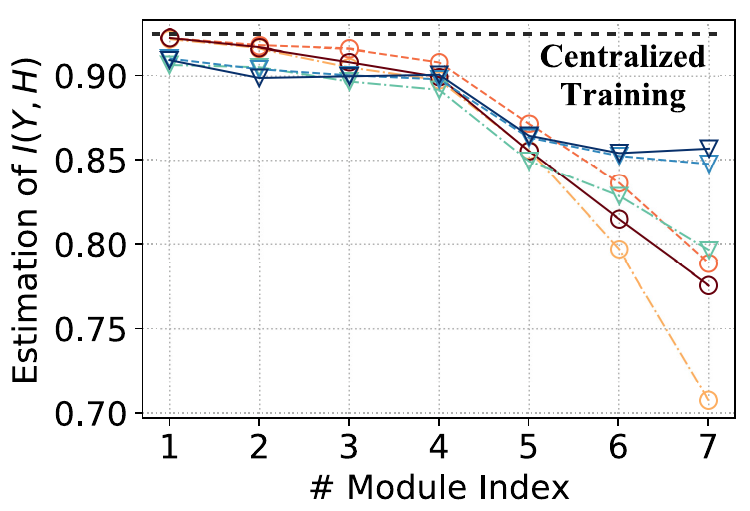}}
     \subfigure[The separability of layers.]{\includegraphics[width=0.3\textwidth]{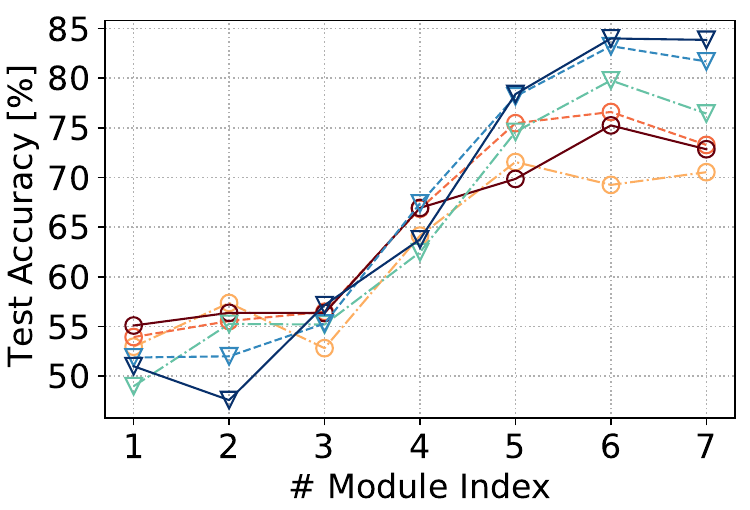}}
     \caption{Estimated MI and separability of trained models with non-IID datasets.}
    \label{fig:MIcompare}
\unishrinkB{}
\end{figure*}

\begin{figure*}[htb!]
    \subfigbottomskip=-1pt
    \subfigcapskip=1pt
  \centering
     \subfigure[Estimated MI $I(H^k;X)$.]{\includegraphics[width=0.3\textwidth]{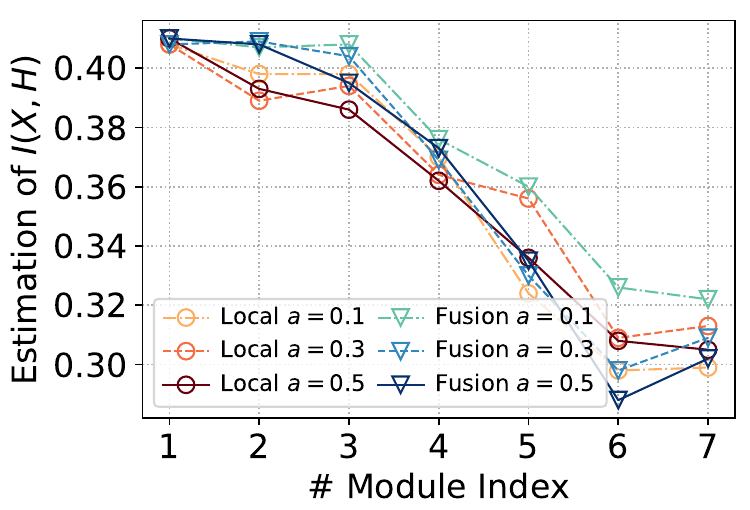}}
     \subfigure[Estimated MI $I(H^k;Y)$.]{\includegraphics[width=0.3\textwidth]{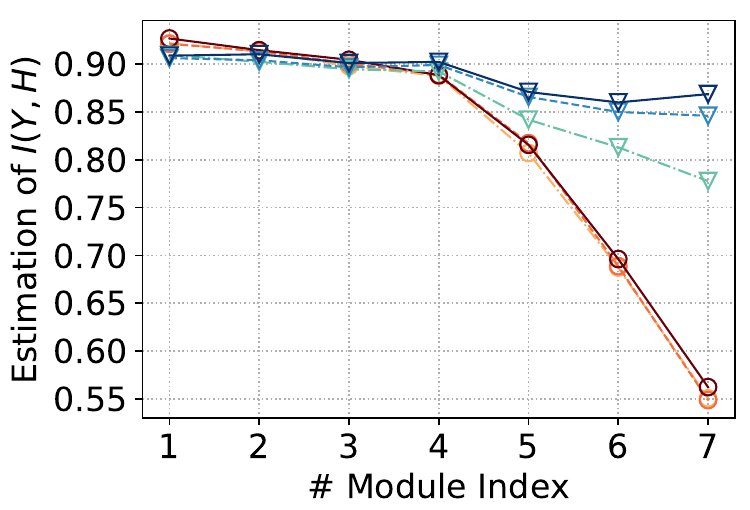}}
     \subfigure[The separability of layers.]{\includegraphics[width=0.3\textwidth]{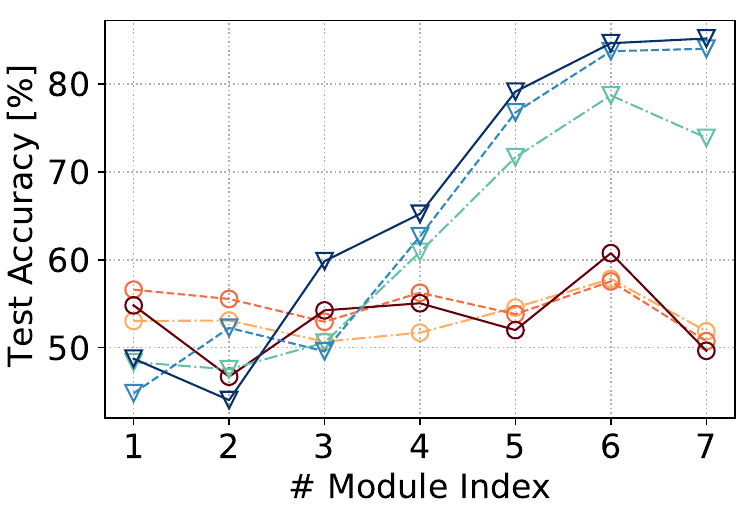}}
     \caption{Estimated MI and separability of trained models with non-IID backdoored datasets.}
    \label{fig:MI-BD-compare}
\unishrinkB{}
\end{figure*}

\textbf{Empirical study.} We empirically estimate the MI $I(H^k_{\tloc}, X)$ and $I(H^k_{\tloc}, Y)$ of isolated local trained features, and the $I(H^k_{\tfus}, X)$ and $I(H^k_{\tfus}, Y)$ of augmented features. As the deep neural networks (DNNs) show layer-wise feature enhancements~\citep{alain2016understanding}, we also measure the linear separability~\citep{alain2016understanding,NIPS2017_10ce03a1} of features $H^k_{\tloc}$ and $H^k_{\tfus}$ to see how they change. Details of MI estimation and linear separability are shown in Appendix~\ref{apdx:EstMI} and ~\ref{apdx:EstLP}. The experiments are conducted by training ResNet-18 with CIFAR-10~\citep{krizhevsky2010cifar} partitioned across $M=5$ clients. Figure~\ref{fig:MIcompare} shows the local features $H^k_{\tloc}$ have significantly higher $I(H^k, X)$ but lower $I(H^k, Y)$ than augmented features $H^k_{\tfus}$. With the increased non-IID degree (lower $a$), the $I(H^k, Y)$ decreased further, demonstrating that the local feature $H^k_{\tloc}$ fits on a more anti-causal relationship between $\SPU{m}$ and $Y_m$. The fused high-level features show better linear separability. And $H^k_{\tfus}$ is more robust to $\SPU{m}$.

Except for the natural spurious features that exist in CIFAR-10, we also study the effect of spurious features by handcraft. Specifically, we inject backdoored data samples~\citep{bagdasaryan2020backdoor,melis2019exploiting} of $1$ out of $5$ clients as $\Dcal_1^{\tback}$, in which the images have handcrafted textures generated according to the labels as a strong anti-causal relation. Details of backdoored datasets are introduced in Appendix~\ref{apdx:backdoor}. Figure~\ref{fig:MI-BD-compare} shows that the backdoor features lead to information loss in $X$. And the backdoored data samples further aggravate the information loss of label $Y$. The isolated local trained features retain significantly less $I(H^k, Y)$ than augmented features.



\unishrinkA{}
\section{FuseFL: Progressive FL Model Fusion}\label{sec:method}
\unishrinkA{}
Motivated by analysis in Section~\ref{sec:analysis}, we propose augmenting local intermediate features $H_m^{1,...,K}$ on client $m$, which helps reduce fitting to a spurious correlation between $Y_m$ and $\SPU{m}$. However, direct fusing features together to make predictions faces the following  problems.

\begin{figure*}[t!]
\unishrinkB{}
\setlength{\belowdisplayskip}{2pt}
    \setlength{\abovedisplayskip}{-5pt}
   \centering
    {\includegraphics[width=0.85\linewidth]{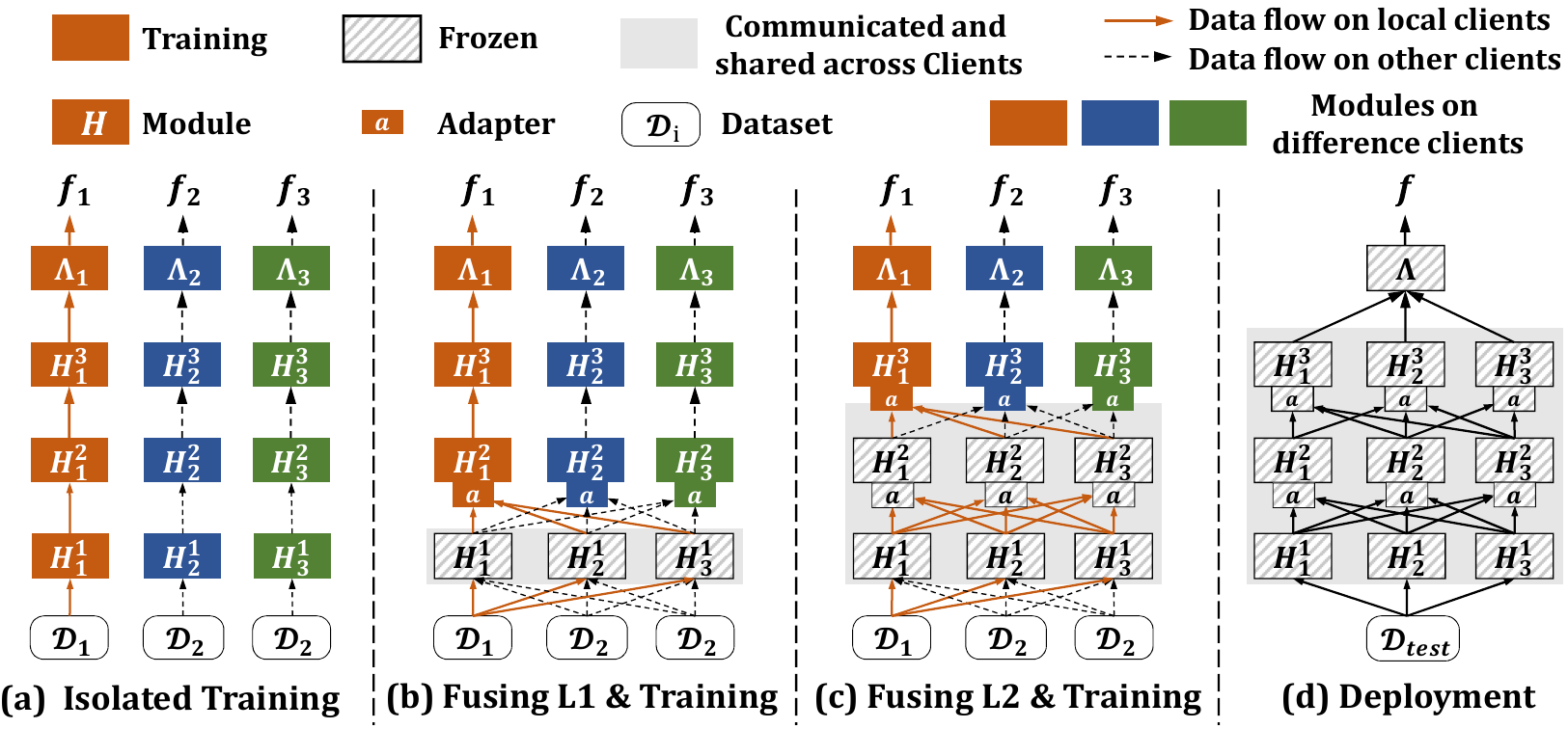}}
    \unishrinkB{}
    \caption{(a) Initially, all layers are isolated training. Note that the layer here does not only mean one or Conv layer, but generally refers to a neural network block that can consist of multiple layers. (b) Then, all first blocks (\textbf{L1}) of different clients are communicated, shared and frozen among clients. Then, the adaptors are added behind the fused block, to fuse features outputted from the concatenated local blocks. (c) Train the third blocks  (\textbf{L3}) follow the similar process in (b).
    (d) inference process of \Ouralgo{}. The larger squares represent the original training block in local models. The smaller squares are adaptors that fuse features from previous modules together, which are $1\times1$ Conv kernels or simple average operations with little or no memory costs. Note that (a) also represents local training in ensemble FL, where different clients train models on local datasets.}
    \label{fig:MainFrameworkTraining}
\unishrinkB{}
\end{figure*}

\textbf{Altered feature distribution.} During local training, the local subsequent model $\Gamma^{k+1}_m \triangleq \Lambda_m \circ H^K_m \cdots \circ H^{k+1}_m $ after block $k$ on client $m$ is trained based on local features $H^k_m$. After feature fusion, changed local features lead to feature drifts~\citep{kumar2022finetuning,VHL}.

\textbf{Mismatched semantics.} Each local feature $H^k_m$ has totally different distributions, scales, or even dimensions; thus, directly averaging may cause useful features to be overwhelmed or confused by noisy features.



\subsection{Train, Fuse, Freeze, and Re-Train}\label{sec:freeze-train}
\unishrinkA{}
As Figure~\ref{fig:MainFrameworkTraining} and Algorithm~\ref{algo} (Appendix~\ref{algo}) show, the main loop of \Ouralgo{} including training, fusing, and freezing, which is repeated for all $K$ split blocks following a progressive manner. Note that for $H^1_m$, there is no layer fusion, which is isolatedly trained. 

\textbf{Fuse.} After each local training step, clients share their $H^k_m$ with other clients, and fuse them following Eq.~\ref{eq:fuse}. An adaptor $A$ is stitched before $H^{k+1}_m$ (Section~\ref{sec:feat-adapt}). Then the local model becomes as $\Ffus{k,m}$ following Eq.~\ref{eq:FuseNN}.

\textbf{Freeze and re-train.} To address the altered feature distribution problem ($A(\Hfus{k}(x)) \to H^k_m(x)$), for each step $k$ on client $m$, the subsequent layers $\Gamma^{k+1}_m$ will be trained again based on $\Dcal_m$. Thus, the $\Gamma^{k+1}_m$ can learn from all low-level features of all clients. Note that we do not need to train each block $k$ with the same epochs in isolated training, because the DNNs naturally follow the layer-wise convergence~\citep{SVCCA,PipeTransformer}.

\begin{align}
\unishrinkB{}
    \Hfus{k}(x) & = \sbr{H^{k}_1(x),...,H^{k}_M(x)}. \label{eq:fuse} \\ 
    \Ffus{k,m} & = \Lambda_m \circ H^K_m \circ ... \circ H^{k+1}_m  \circ  \Hfus{k} \circ ... \circ \Hfus{1}. \label{eq:FuseNN} \\
    \Ffus{} & = \Lambda \circ \Hfus{K} \circ \Hfus{K-1} \circ ... \circ \Hfus{1}. \label{eq:FinalFuseF} \\
    A_{\text{avg}} & = \text{Average}(\Hfus{k}(x)). \label{eq:AVGada} \\ 
    A_{\text{conv}} & = \text{Conv}(\text{Concat}(\Hfus{k}(x))). \label{eq:ConvAda}
\unishrinkB{}
\end{align}

By freezing fused blocks and retraining high-level models, the another benefit is to enforce the SGD to use previous features from other clients to continue tuning the high-level model. The previous local trained high-level models may overfit on shortcut features from the noisy data. This insight is also utilized in defensing adversarial attacks~\citep{utrera2020adversarially,papernot2018deep}.




\subsection{Feature Adaptation}\label{sec:feat-adapt}
\unishrinkA{}
To address the mismatched semantics problem, the intuitive approach is to preserve the original feature structures through the concatenation of all features to the next block. However, this leads to new problems: (1) requiring modification of subsequent modules; (2) feature size explosion of subsequent blocks by $O(M^K)$. To address these two problems, we introduce an adaptor stitched before local modules ($k > 1$), and training together with $\Gamma_m^{k}$. As an initial trial to operationalize \Ouralgo{}, we utilize \conv{} as the adapter as Eq.~\ref{eq:ConvAda}. We also verify the use of average as an adapter (Eq.~\ref{eq:AVGada}) in experiments (Section~\ref{sec:exp}).



\unishrinkA{}
\subsection{Benefits of \Ouralgo{} Design}\label{sec:method-benefits}
\unishrinkA{}
\textbf{Mitigating fitting on spurious correlations.}
During the local training on datasets with spurious features, the final learned representations with ERM still contain some invariant features~\citep{ahuja2021invariance,chen2023understanding}. Thus, some work proposes to finetune the classifier based on data samples with invariant features to let the classifier make predictions based on invariant features~\citep{kirichenko2023last,izmailov2022feature,pagliardini2023agree}. Similar to this motivation, we hope to incorporate other client modules as auxiliary feature extractors to generate more invariant features of local data during training subsequent layers. Figure~\ref{fig:Motivation} describes the mechanism that using other local models $H_{i\neq m}(X_m)$ help to filter out spurious features $R_{m}^{\text{spu}}$, but retain $R_{m}^{\text{inv}}$. As other clients $\left\{i\neq m\right\}$ cannot see $R_{m}^{\text{spu}}$ in their dataset $\mathcal{D}_{i \neq m}$ during local training, only invariant features can pass through $H_{i\neq m}(X_m)$. This method can be seen as the invariant feature augmentation~\citep{chen2023understanding}.

\textbf{Saving storage and communication costs than ensemble FL.} Similar to ensemble learning, directly collecting and fusing local models together will enlarge the total model size from $\size$ to $\size \times M$. Note that \Ouralgo{} actually builds up a global model with blocks fused together, with the hidden dimensions (channels) enlarged from $n_s \to n_s\times M $. Thus, intuitively, we can reduce the hidden dimension of the local model $n_f $ to reduce the memory requirements. Interestingly, with a scaling ratio $\gamma$, when scaling all local linear or convolutional layers, each matrix should be scaled on \textit{both input and output} dimension as $n_f = \gamma n_s$. The ratio of memory costs between \Ouralgo{} and the original single model is $r_m = M\times n_s^2 / (\gamma n_s)^2$. To obtain $r_m=1$, we obtain the scaling ratio $\gamma=\sqrt{M}$, which means that \Ouralgo{} can keep similar memory requirements with the original model size $\size$ with reducing hidden dimensions as ratio $\sqrt{M}$, demonstrating good theoretical scalability to the $M$. We will verify this in experiments (Section~\ref{sec:exp_results}).


\textbf{Privacy concerns.} \Ouralgo{} only shares layers between clients, which aligns with other classic and advanced FL methods in all directions mentioned in Section~\ref{sec:related}.

\textbf{Support of heterogeneous models.} The block in \Ouralgo{} does not mean a single linear or convolution layer, but a general module that can consist of any DNN, thus supporting FL with heterogeneous models (see experiments~\ref{sec:exp_results}). The adaptor can be designed to transform features of different shapes to align with the input of the next local block.

\textbf{Layer-wise training to reduce training epochs.} Because each communication round means multiple local training epochs. To keep the total training epochs the same as the one-shot FedAVG (represented as $E$), we assign the local training epochs of the \Ouralgo{} as $E/K$. Thus, the number of total training epochs of FuseFL is the same as other OFL methods. The core insight of this design can be referred to as the progressive freezing during training DNNs ~\citep{SVCCA}.



\section{Related Works}\label{sec:related}


\subsection{Data Heterogeneity in FL}\label{sec:relatedNoniid}
\begin{wrapfigure}{R}{0.5\textwidth}
\begin{minipage}{0.5\textwidth}
\begin{table}[H]
\centering
\caption{Demystifying different FL algorithms. $T$ represents communication rounds, $S$ the model size, $M$ the number of clients. The ``Centralized'' means training the model with all datasets aggregated with SGD. ``Comm.'' means communication. } 
\vspace{-0.0cm}
\centering
\begin{adjustbox}{max width=\linewidth}
\begin{tabular}{c|cccc}
\toprule[1.5pt]
Method & \makecell{Comm. \\ cost} & Storage cost & \makecell{Support model\\ heterogeneity} & \makecell{Not require \\extra data}  \\
\midrule[1.5pt]
FedAvg &  $T\times S $ & $S$  &  \cmark  &  \cmark \\
\midrule[1pt]
Average-based OFL & $S$  & $S$ &\xmark  & \xmark  \\
Ensemble-based OFL &  $S$ & $S\times M$ &  \cmark  &  \xmark \\
Model distillation OFL &  $S$ &  $S$ &  \cmark & \xmark  \\
\midrule[1pt]
\Ouralgo{} &  $S$  & $S$ & \cmark & \cmark  \\
\bottomrule[1.5pt] 
\end{tabular}
\end{adjustbox}
\vspace{-0.5cm}
\label{tab:demystify}
\end{table}
\end{minipage}
\end{wrapfigure}
The notorious non-IID data distribution in FL severely harms the convergence rate and model performance of FL. The \textbf{model regularization} proposes to add a penalty of distances between local and global models~\citep{fedprox,acar2021federated}. \textbf{Feature calibration} aligns feature representations of different clients in similar spaces~\citep{dong2022spherefed,VHL,10203389,tang2024fedimpro}. FedMA~\citep{wang2020Federated} exploits a layer-wise communication and averaging methods, in which the aggregation is conducted on fine-grained layer. Thus, its linear dependence of computation and communication on the network’s depth, is not suitable for deeper models, which introduces large computation costs in re-training and more communication rounds~\citep{aggregation2,SixuNeuronMatch}. Unlike FedMA, FuseFL introduces block-wise communication and aggregation with much less communication rounds and computation costs. Furthermore, due to the matching and averaging aggregation, FedMA only supports linear or Conv layers, which severely limits its practical usage. By viewing the separated block as a black box and concatenating output features, FuseFL can successfully support merging any kind of neural layer. Some works on fairness analysis in FL also relate to this work in perspective of local and global characteristics~\citep{hamman2024demystifying,ezzeldin2023fairfed}.

\subsection{One-shot FL}\label{sec:relatedOneshot}
One-shot FL~\citep{zhang2022dense,guha2019one,li2020practical,zhou2020distilled,dennis2021heterogeneity} reduces communication costs from $T\times\size $ to $\size$ by communicating with only one round. \textbf{Average-based} methods focus on better averaging client models, like Fisher information~\citep{jhunjhunwala2023towards,qu2022generalized}, bayesian optimization~\citep{liu2023bayesian,al2020federated} or matching neurons~\citep{ainsworth2022git, wang2020Federated}. However, the non-linear structure of DNNs makes it difficult to obtain a comparable global model through averaging. \textbf{Ensemble-based} methods make prediction based on all or selected client models~\citep{diao2023towards,heinbaugh2023datafree,wang2023datafree}, but requires additional datasets with privacy concerns. And they have low scalability of the number of clients due to the storage of client models. \textbf{Model distillation} uses the public~\citep{li2020practical} or synthesized datasets~\citep{zhou2020distilled,dennis2021heterogeneity} to distill a new model based on ensemble models~\citep{guha2019one,li2020practical}. These methods may be impractical in data-sensitive scenarios, such as medical and education, or continuous learning scenarios~\citep{dong2022federated_FCIL,10323204}. Furthermore, there exists a large performance gap between these methods and the ensemble learning.

Due to the limited space, we leave the detailed reviews in Table~\ref{tab:demystify-detail}  and Appendix~\ref{apdx:related}. Table~\ref{tab:demystify} concisely demystifies different FL methods in terms of communication cost, storage cost, model performance, and whether supporting model heterogeneity or requiring extra data.

\section{Experiments}\label{sec:exp}

\subsection{Experiment Setup}

\textbf{Federated Datasets and Models.}
In order to validate the efficacy of \Ouralgo{}, we conduct comprehensive experiments with commonly used datasets in FL, including MNIST~\citep{726791}, CIFAR-10~\citep{Krizhevsky09learningmultiple}, FMNIST~\citep{xiao2017fashion}, SVHN~\citep{SVHN},  CIFAR-100~\citep{Krizhevsky09learningmultiple} and Tiny-Imagenet~\citep{le2015tiny}. For studying the non-IID problem in FL, we partitioned the datasets through a widely-used non-IID partition method, namely Latent Dirichlet Sampling~\citep{LDA,kairouz2021advances,reddi2020adaptive,li2021federatedstudy}, in which the coefficient $a$ represents the non-IID degree. Lower $a$ generates more non-IID datasets, and vice versa. Consistent with established practices in the field~\citep{zhang2022dense,reddi2020adaptive, luo2021no,li2021federatedstudy}, each dataset was divided with three distinct degrees of non-IID with $a \in \cbr{0.1, 0.3, 0.5}$. If there is no additional explanation, the non-IID degree $a$ is set to $0.5$ by default.

Following other classic and advanced FL works studying non-IID problems and communication-efficient FL~\citep{sun2023fedspeed,heinbaugh2023datafree,luo2021no,zhang2022dense}, we train ResNet-18~\citep{resnet} on all datasets in main experiments. And we reduce and increase the number of layers as ResNet-10 and ResNet-26 to verify the effect of \Ouralgo{} in model-heterogeneity FL. The number of clients is set as $M=5$ by default. Moreover, we study the scalability of our methods with different $M\in \cbr{5, 10, 20, 50}$.

We use SGD optimizer with momentum coefficient as 0.9, and the batch size is 128. The number of local training epochs $E=200$. We search learning rates in $\left \{0.0001, 0.001, 0.01, 0.1 \right \}$ and report the best results. The detailed hyper-parameters of different settings are shown in Table~\ref{tab:LearningRate} of Appendix~\ref{apdx:DetailedExp}.

\begin{table*}[htb!]
\centering
\caption{Accuracy of different 
methods across $\alpha=\{0.1, 0.3, 0.5\}$ on different datasets. Ensemble means ensemble learning with local trained models, which is an upper bound of all previous methods but impractical in FL due to the large memory costs and the weak scalability of clients. Thus, we highlight the best results in \textbf{bold font} except Ensemble.}
\vspace{-0.2cm}
\begin{adjustbox}{max width=\linewidth}
\begin{tabular}{c|ccc|ccc|ccc|ccc|ccc|ccc}
\toprule
Dataset & \multicolumn{3}{c|}{MNIST} & \multicolumn{3}{c|}{FMNIST} & \multicolumn{3}{c|}{CIFAR-10} & \multicolumn{3}{c|}{SVHN} & \multicolumn{3}{c}{CIFAR-100} & \multicolumn{3}{c}{Tiny-Imagenet} \\
\midrule
Method & $\alpha$=0.1 & $\alpha$=0.3 & $\alpha$=0.5 & $\alpha$=0.1 & $\alpha$=0.3 & $\alpha$=0.5 & $\alpha$=0.1 & $\alpha$=0.3 & $\alpha$=0.5 & $\alpha$=0.1 & $\alpha$=0.3 & $\alpha$=0.5 & $\alpha$=0.1 & $\alpha$=0.3 & $\alpha$=0.5  & $\alpha$=0.1 & $\alpha$=0.3 & $\alpha$=0.5 \\
\midrule
FedAvg & 48.24 & 72.94 & 90.55 & 41.69 & 82.96 & 83.72 & 23.93 & 27.72 & 43.67 & 31.65 & 61.51 & 56.09 & 4.58 & 11.61 & 12.11 &  3.12 & 10.46 & 11.89 \\
FedDF & 60.15 & 74.01 & 92.18 & 43.58 & 80.67 & 84.67 & 40.58 & 46.78 & 53.56 & 49.13 & 73.34 & 73.98 & 28.17 & 30.28 & 36.35 & 15.34 & 18.22 & 27.43   \\
Fed-DAFL & 64.38 & 74.18 & 93.01 & 47.14 & 80.59 & 84.02 & 47.34 & 53.89 & 58.59 & 53.23 & 76.56 & 78.03 & 28.89 & 34.89 & 38.19 & 18.38 & 22.18 & 28.22 \\
Fed-ADI & 64.13 & 75.03 & 93.49 & 48.49 & 81.15 & 84.19 & 48.59 & 54.68 & 59.34 & 53.45 & 77.45 & 78.85 & 30.13 & 35.18 & 40.28  &  19.59 & 25.34 & 30.21 \\
DENSE & 66.61 & 76.48 & 95.82 & 50.29 & 83.96 & 85.94 & 50.26 & 59.76 & 62.19 & 55.34 & 79.59 & 80.03 & 32.03 & 37.32 & 42.07 & 22.44 & 28.14 & 32.34 \\
\midrule
\rowcolor{greyL} Ensemble &86.81          & 96.76          & 97.22          & 67.71          & 87.25          & 89.42                        & 57.5           & 77.35          & 79.91          & 65.29          & 88.31          & 85.7           & 35.69          & 49.41          & 53.39          &  30.85 & 39.43 & 45.8   \\
\midrule
\Ouralgo{} $K=2$ & 97.02          & \textbf{98.43}          & \textbf{98.54}          & 83.15          & \textbf{89.94}          & 89.47                        & 70.85          & 81.41          & \textbf{84.34}          & 76.88          & \textbf{91.07}          & \textbf{90.87}          & 34.07          & \textbf{45.12}          & 46.12          & \textbf{29.28} & 31.11 & \textbf{34.34} \\
\Ouralgo{} $K=4$ & \textbf{97.19}          & 98.34          & 98.29          & 83.05          & 84.58          & \textbf{90.50}           & \textbf{73.79}          & \textbf{84.58}          & 81.15          & 78.08          & 89.63          & 89.34          & \textbf{36.86}          & 42.79          & \textbf{49.30}           & 27.63 & \textbf{33.04} & 34.28  \\
\Ouralgo{} $K=8$ & 96.66          & 98.35          & 98.16          & \textbf{83.2}           & 88.57          & 88.24          & 70.46          & 80.70           & 74.99          & \textbf{80.31}          & 88.88          & 89.94          & 34.97          & 39.08          & 40.73         & 25.21 & 32.59 & 33.82 \\
\bottomrule
\end{tabular}
\end{adjustbox}
\label{tab:main}
\vspace{-0.0cm}
\end{table*}
\textbf{Baselines.} 
Except the classic baseline FedAvg~\citep{mcmahan2017communication} and advanced OFL method DENSE~\citep{zhang2022dense}, we apply two prevailing data-free KD methods DAFL~\citep{chen2019data} and ADI~\citep{yin2020dreaming} into OFL. We choose FedDF~\citep{FedDF} as its high efficiency in few-round FL. We conduct ensemble FL as it is the upper bound across ensemble-and-distillation based methods, yet impractical in real-world scenarios. The communication round for all baseline methods is only 1. For our method \Ouralgo{}, the number of communication rounds is equal to the number of splitted blocks $K$. However, the actual communication cost is as same as one-shot FL. Because \Ouralgo{} only communicates a part of the model. After all rounds, the total communication cost is $SM$, where $S$ is the model size, $M$ the number of clients.

\subsection{Experimental Results}\label{sec:exp_results}

\textbf{Main Results.} Table~\ref{tab:main} shows that \Ouralgo{} generally outperforms all other baselines except for ensemble FL. All ensemble-and-distillation baselines  have lower performance than ensemble FL. Nevertheless, by the insights from causality (Section~\ref{sec:analysis}) and our innovative design (Section~\ref{sec:method}), \Ouralgo{} can significantly outperform ensemble FL for almost all cases except for CIFAR-100 with $a=0.3, 0.5$ and Tiny-Imagenet. We suppose the reason is that CIFAR-100 and Tiny-Imagenet has much more data divergence between different classes, thus the overlap between $\INV{m}$ is much less than other datasets. Recall that the benefits of FuseFL come from fusing sub models training on other clients, thus filtering out $R_{m}^{\text{spu}}$ and collecting $R_{m}^{\text{inv}}$ of the same class to improve the generalization performance. The large data divergence in CIFAR-100 and Tiny-Imagenet limits benefits of \Ouralgo{}.

\begin{minipage}{\textwidth}
 \begin{minipage}[t]{0.45\textwidth}
  \centering
     \makeatletter\def\@captype{table}\makeatother\caption{Accuracy with \Ouralgo{} with \conv{} or averaging to support heterogeneous model design on CIFAR-10. }
    \vspace{-0.05cm}
    \begin{adjustbox}{max width=\linewidth}
    \begin{tabular}{c|ccc}
    \toprule
    non-IID degree & $a=0.1$        & $a=0.3$        & $a=0.5$        \\
    \midrule
    \rowcolor{greyL} Ensemble               & 57.5           & 77.35          & 79.91          \\
    \midrule
    \Ouralgo{}              & \textbf{73.79} & \textbf{84.58} & 81.15   \\
    \Ouralgo{} (Avg)        & 68.08          & 71.49          & 80.35   \\
    \Ouralgo{}-Hetero       & 75.33          & 81.71          & \textbf{82.71}   \\
    \Ouralgo{} (Avg)-Hetero & 68.31          & 76.27          & 79.74  \\
    \bottomrule
    \end{tabular}
    \end{adjustbox}
    \label{tab:modelHetero}
  \end{minipage}
  \begin{minipage}[t]{0.54\textwidth}
   \centering
        \makeatletter\def\@captype{table}\makeatother\caption{Memory Occupation. For different number of clients, the number of basic channels in ResNet-18 of \Ouralgo{} is set as 32, 20, 14, 9 with $M\in\cbr{5, 10, 20, 50}$, respectively. Other OFLs refer to FedAvg, FedDF, Fed-DAFL, Fed-ADI, DENSE.}
        \begin{adjustbox}{max width=\linewidth}
        \begin{tabular}{c|cccc}
        \toprule
        \# Clients        & $M=5$    & $M=10$   & $M=20$   & $M=50$    \\
        \midrule
        \rowcolor{greyL} Single Model     & 42.66MB  & 42.66MB  & 42.66MB  & 42.66MB   \\
        \midrule
        Other OFLs	 & 42.66MB & 42.66MB & 42.66MB & 42.66MB \\
        Ensemble         & 213.31MB & 426.62MB & 853.24MB & 2133.10MB \\
        \Ouralgo{} $K=2$ & 53.71MB  & 42.32MB  & 42.13MB  & 45.48MB   \\
        \Ouralgo{} $K=4$ & 55.38MB  & 44.92MB  & 47.22MB  & 58.63MB   \\
        \Ouralgo{} $K=8$ & 68.08MB  & 64.77MB  & 86.11MB  & 159.08MB  \\
        \Ouralgo{} (Avg)      & \textbf{53.32MB}  & \textbf{41.66MB}  & \textbf{40.83MB}  & \textbf{42.18MB}  \\
        \bottomrule
        \end{tabular}
        \label{tab:memory}
        \end{adjustbox}
   \end{minipage}
\end{minipage}

\textbf{Support of heterogeneous model design.}
Table~\ref{tab:modelHetero} shows training heterogeneous model using \Ouralgo{}. In all $M=5$ clients, 2 clients train ResNet-10 and other 2 clients train ResNet-26, the left 1 client trains ResNet-18. We set $K=4$ for \Ouralgo{}. Results show that training with heterogeneous models has similar or even better results than homogeneous models, demonstrating that \Ouralgo{} supports training heterogeneous model well. This will be very useful in heterogeneous computation environments.

\begin{table}[htb!]
\centering
\caption{Accuracy of different 
methods across $M=\{5, 10, 20, 50\}$. }
\vspace{-0.05cm}
\begin{adjustbox}{max width=0.85\linewidth}
\begin{tabular}{c|cccc|cccc}
\toprule
Dataset            & \multicolumn{4}{c}{CIFAR-10}      & \multicolumn{4}{c}{SVHN}      \\
\toprule
Method             & $M=5$ & $M=10$ & $M=20$ & $M=50$ & $M=5$ & $M=10$ & $M=20$ & $M=50$ \\
\midrule
FedAvg             & 43.67 & 38.29  & 36.03  & 23.01  & 56.09 & 45.34  & 47.79  & 36.53  \\
FedDF              & 53.56 & 54.44  & 43.15  & 29.52  & 73.98 & 62.12  & 60.45  & 51.44  \\
Fed-DAFL           & 55.46 & 56.34  & 45.98  & 29.41  & 78.03 & 63.34  & 62.19  & 54.23  \\
Fed-ADI            & 58.59 & 57.13  & 46.45  & 27.45  & 78.85 & 65.45  & 63.98  & 57.35  \\
DENSE              & 62.19 & 61.42  & 52.71  & 28.51  & 80.03 & 67.57  & 66.42  & 59.27  \\
\midrule
\rowcolor{greyL} Ensemble           & 79.91 & 77.25  & 59.69  & 55.63  & 85.7  & 73.45  & 68.76  & 54.96  \\
\midrule
\Ouralgo{} $K=2$ & \textbf{84.34} & 73.71  & 62.85  & \textbf{42.18}  & \textbf{90.87} & 88.52  & 85.18  & \textbf{72.25}  \\
\Ouralgo{} $K=4$ & 81.15 & \textbf{78.28}  & 62.57  & 37.08  & 89.34 & \textbf{89.31}  & \textbf{86.94}  & 45.49  \\
\Ouralgo{} $K=8$ & 74.99 & 67.35  & \textbf{63.19}  & 28.28   & 89.94 & 72.65  & 64.11  &  42.19   \\
\bottomrule
\end{tabular}
\end{adjustbox}
\vspace{-0.0cm}
\label{tab:scalability}
\end{table}


\begin{wrapfigure}{R}{0.5\textwidth}
\begin{minipage}{0.5\textwidth}
\begin{table}[H]
\centering
\caption{Local and global accuracy of 5 local client models. BD$_0$ and BD$_1$ represent two clients trained on backdoored datasets. Normal$_0$, Normal$_1$, and Normal$_2$ represent three clients trained on clean datasets. }
\vspace{-0.0cm}
\centering
\begin{adjustbox}{max width=\linewidth}
\begin{tabular}{c|ccccc}
\toprule[1.5pt]
Client & BD$_0$ & BD$_1$ & Normal$_0$ & Normal$_1$ & Normal$_2$ \\
\midrule[1.5pt]
Local Acc. & 100.0 & 100.0 & 99.7 & 99.9 & 100.0 \\
Global Acc. & 32.6 & 27.1 & 41.2 & 42.3 & 38.4 \\
\bottomrule[1.5pt] 
\end{tabular}
\end{adjustbox}
\vspace{-0.5cm}
\label{tab:BD-acc}
\end{table}
\end{minipage}
\end{wrapfigure}

\textbf{Memory occupation.} Table~\ref{tab:memory} shows the memory occupation with varying number of clients and split modules. Results show that \Ouralgo{} requires similar memory with the single model when $K=2$, while the ensemble FL requires the $S\times M$ storage cost. And using averaging as feature fusion method can further reduce the memory cost. Table~\ref{tab:conv1x1VSavg} shows that the \Ouralgo{} with $K=2$ or \Ouralgo{} with averaging has comparable or better performance than other variants of \Ouralgo{}.

\begin{wrapfigure}{R}{0.5\textwidth}
\begin{minipage}{0.5\textwidth}
\begin{table}[H]
\centering
\caption{Comparing accuracy on backdoored CIFAR-10. }
\begin{adjustbox}{max width=\linewidth}
\begin{tabular}{c|ccc}
\toprule
\rowcolor{greyL} \# Backdoored clients & \multicolumn{3}{c}{$M_{bd}=1$} \\
\midrule
Non-IID degree   & $a=0.1$  & $a=0.3$  & $a=0.5$  \\
\midrule
Ensemble         & 52.76    & 74.88    & 73.46    \\
\Ouralgo{} $K=2$ & 43.99    & 75.61    & \textbf{84.79}    \\
\Ouralgo{} $K=4$ & \textbf{55.99}    & \textbf{78.39}    & 80.52    \\
\Ouralgo{} $K=8$ & 45.66    & 74.01    & 82.12    \\
\bottomrule
\rowcolor{greyL} \# Backdoored clients & \multicolumn{3}{c}{$M_{bd}=2$} \\
\midrule
Non-IID degree        & $a=0.1$  & $a=0.3$  & $a=0.5$  \\
\midrule
Ensemble         & 52.67    & 68.42    & 75.83    \\
\Ouralgo{} $K=2$ & 54.24    & \textbf{71.49}    & \textbf{82.61}    \\
\Ouralgo{} $K=4$ & \textbf{54.77}    & 67.53    & 72.99    \\
\Ouralgo{} $K=8$ & 51.70     & 68.31    & 76.95   \\
\bottomrule
\end{tabular}
\label{tab:backdooracc}
\end{adjustbox}
\end{table}
\end{minipage}
\end{wrapfigure}

\textbf{Scalability of the number of clients.} We empirically prove that the memory occupation increases a little along with the increased $M$ (Table~\ref{tab:memory}), and keeping performance outperforms than baselines (Table~\ref{tab:scalability}).

\textbf{Influence on local models of backdoored datasets.} As shown in Section~\ref{sec:analysis}, the backdoored features lead to information loss in $X$. Here we further show how the FL performance is influenced by the backdoored features. Table~\ref{tab:BD-acc} shows that the backdoored (BD) clients fit on the handcrafted spurious features, thus having lower global accuracy than normal clients.

\textbf{Test accuracy on backdoored datasets.}
Table~\ref{tab:backdooracc} provides the test accuracy of different methods training with backdoored CIFAR-10. The test dataset is the original clean test set. We test different numbers of backdoored clients $M_{bd} = 1,2$ out of a total of 5 clients, to see how the degree of the backdoor influences training. Results show \Ouralgo{} outperforms ensemble FL in all cases. demonstrating that \Ouralgo{} can defend better against the backdoor samples than ensemble FL.

\section{Conclusion}\label{sec:conclusion}
In this work, we draw inspiration from the causality and the information bottleneck to analyze the cause of low performance of ensemble FL and OFL. Specifically, the local isolatedly trained models are easily to fit spurious features, as local clients cannot learn more invariant features and remove spurious features from other datasets. Built upon this insight, we provide a novel approach \Ouralgo{} to augment features by fusing client layers in a bottom-up manner, thus mitigating the spurious fitting and encourage learning of invariant features. \Ouralgo{} achieves OFL with extremely low communication costs with significantly higher performance than current OFL and ensemble FL methods.

\begin{ack}
This work was partially supported by National Natural Science Foundation of China under Grant No. 62272122, the Guangzhou Municipal Joint Funding Project with Universities and Enterprises under Grant No. 2024A03J0616, the Hong Kong RIF grant under Grant No. R6021-20, and Hong Kong CRF grants under Grant No. C2004-21G and C7004-22G. BH was supported by NSFC General Program No. 62376235, Guangdong Basic and Applied Basic Research Foundation Nos. 2022A1515011652 and 2024A1515012399, HKBU Faculty Niche Research Areas No. RC-FNRA-IG/22-23/SCI/04, and HKBU CSD Departmental Incentive Scheme. YGZ and YMC were supported in part by the NSFC / Research Grants Council (RGC) Joint Research Scheme under the grant: N-HKBU214/21, the General Research Fund of RGC under the grants: 12201321, 12202622, 12201323, the RGC Senior Research Fellow Scheme under the grant: SRFS2324-2S02, and the Initiation Grant for Faculty Niche Research Areas of Hong Kong Baptist University under the grant: RC-FNRA-IG/23-24/SCI/02.
\end{ack}


\bibliography{cite}



\newpage
\appendix

\section*{Appendix / supplemental material}
\section{Details of Algorithm}\label{apdx:algo}
Details of our algorithm is shown in Algorithm~\ref{algo}.

\begin{algorithm}[htp]
	\caption{FL with \Ouralgo{}}
	\label{algo}
	\textbf{Input: } The number of split modules $K$; Initialized local modules $H_{m}^1,...,H_{m}^K$ and classifier $\Lambda_m$.\\
    \textbf{Output:} The fused model $\Ffus{}$ (Eq.~\ref{eq:FinalFuseF}). \\
    \begin{algorithmic}[1]
        \STATE \textit{\# Train and fuse}
        \FOR{module $ k=1, \cdots, K$}
            \FOR{ each client $ m \in \Mcal$ \textbf{ in parallel do}}
                   \STATE $ \theta^{k}_m \leftarrow$ \texttt{ClientTrain}$(k)$;
            \ENDFOR
            \STATE Fuse $ \Hfus{k} \leftarrow (H^{k}_1,...,H^{k}_M)$ as Eq.~\ref{eq:fuse}; 
            \FOR{ each client $ m \in \Mcal$ \textbf{ in parallel do}}
                   \STATE \texttt{ClientAdapt}$(k, \Hfus{k})$;
            \ENDFOR
        \ENDFOR
        \STATE \textit{\# Calibrate classifier}
        \STATE Averaging and calibrating final classifier $\Lambda$ using CCVR~\citep{luo2021no} with object $\ell(\Ffus{};x,y)$ (Eq.~\ref{eq:FinalFuseF});
        \STATE \textbf{Return} $\Ffus{}$.
       \STATE
        \STATE \texttt{ClientTrain}($ k$):
            \STATE  Build $\Ffus{k,m}$ follows Eq.~\ref{eq:FuseNN};
            \STATE  Freeze $H^{k-1}_m,...,H^1_m$;
            \FOR{each local iteration $j=0,\cdots, E$}
              \STATE $ \theta_{m,j+1} \leftarrow \theta_{m,j} - \eta_{k,j} \nabla_{{\theta}}  \ell(\Ffus{k,m};x,y), x,y\sim\Dcal_m$;
            \ENDFOR
        \STATE \textbf{Return} $H_m^k$ parameterized with $\theta_{m,E}^k$ ;
        \STATE
        \STATE \texttt{ClientAdapt}($ k, \Hfus{k}$):
              \STATE Build adaptor $A^{k+1}_m$ following Eq~\ref{eq:AVGada} or ~\ref{eq:ConvAda};
              \STATE Stitch $H^{k+1}_m \triangleq H^{k+1}_m \circ A^{k+1}_m$ ;
        \STATE
\end{algorithmic}
\vspace{0.1cm}
\end{algorithm}

\section{Broader Impact}\label{apdx:impact}
This work provides a novel OFL approach, aiming at advancing the field of federated learning. There are many potential societal consequences of our work. For instance, our work can extremely reduce the communication cost of FL, reducing energy consumption. Under the low-bandwidth communication environments like Internet with 1$\sim$ 10 MB/s~\citep{tang2023fusionai,yuandecentralized,tang2024fusionllmdecentralizedllmtraining}, this method provides possibility to train super large models like large language models with low communication time. Furthermore, to the best of our knowledge, this is the first work that understands FL from the view of causality. There exist a large space for future works to study along this direction.

\Ouralgo{} can further inspire more research and applications in the communication-constrained scenarios, like extremely low communication bandwidth, and training large language models (LLM) in FL~\citep{tang2024fusionllmdecentralizedllmtraining}. We will try to extend \Ouralgo{} into training LLMs or MoE in FL scenarios with its low communication costs.

To deploy transformer-based frameworks like current LLMs with the core idea of \Ouralgo{} in one-shot FL, we envision two methods here.
\begin{itemize}[leftmargin=2em]
\item \textbf{Concat-and-freeze.} Similar to training ResNet in \Ouralgo{}, we can block-wisely train and collect the transformer blocks together for each round; during local training, the output features of all transformer blocks are concatenated to feed into the subsequent layers. Due to the large resource consumption of pretraining, we do not evaluate this idea here.
\item \textbf{Averaging-and-freeze LoRA.} Here, we consider the finetuning scenarios with LoRA~\citep{huang2024lorahub}. LoRA blocks can be seen as additional matrix mapping applied on the local Q V attentions and MLP layers. The output is the original feature plus the LoRA output. To use LoRA in \Ouralgo{}, we can follow the MoE style~\citep{loramoe} or the averaging style~\citep{huang2024lorahub}. Specifically, we consider averaging LoRAs on different clients together, then averaging and freezing all LoRAs in each transformer block to freeze the obtained aggregated features in each communication round.
\end{itemize}

\section{More related works}\label{apdx:related}
We provide the comprehensive introduction of the related works in this section. The data heterogeneity problem in FL is introduced in Section~\ref{apdx:relatedNoniid}, and the communication compression in FL is introduced in Section~\ref{apdx:compressFL}. Further, by seeing local datasets from other clients as the OOD datasets with respect to the isolated locally trained models, we review some methods in OOD generalization (Section~\ref{apdx:OOD-gene} and mutual information (Section~\ref{apdx:relatedMI}) to build a connection between them and FL with extremely low communication costs. We demystify our work from current works in Table~\ref{tab:demystify-detail}, which is a detailed version of Table~\ref{tab:demystify}.

\begin{table*}[htb!]
\centering
\caption{Demystifying different FL algorithms. $T$ represents communication rounds, $S$ the model size, $M$ the number of clients. Practically, in communication-compression FL, the minimal sparsification ratio $Q_{\text{Spar}}$ is 0.1, the quantization ratio $Q_{\text{Quant}}$ is $0.125 \ (32\to 4 \text{bits})$, and the low rank ratio $Q_{\text{LowR}}$ is 0.1, otherwise the convergence is difficult to achieve and significantly harming model performance. The ``Centralized'' means training the model with all datasets aggregated with SGD. Due to the data heterogeneity, the performance usually is: \textit{Centralized} $\ge$ \textit{FedAvg} $\ge$ \textit{Ensemble}.} 
\vspace{-0.1cm}
\centering
\begin{adjustbox}{max width=\linewidth}
\begin{tabular}{cc|ccccc}
\toprule[1.5pt]
Method type &  Core technique   & \makecell{Communication \\ cost} & Storage cost & \makecell{Performance \\ upper bound} & \makecell{Support model\\ heterogeneity} & \makecell{Not require \\extra data}  \\
\midrule[1.5pt]
\multirow{4}{*}{FedAvg } &  No compression & $T\times S $ & $S$  &  Centralized &  \cmark  &  \cmark \\
&  Sparsification & $T\times S\times Q_{\text{Spar}} $ & $S$  & FedAvg  &  \cmark  &  \cmark \\
&  Quantization & $T\times S\times Q_{\text{Quant}} $  & $S$  & FedAvg & \cmark   &  \cmark \\
&  Low-rank & $T\times S\times Q_{\text{LowR}} $  & $S$  & FedAvg &  \cmark  &  \cmark \\
\midrule[1pt]
\multirow{3}{*}{One-shot FL} &  Average-based & $S$  & $S$ & Ensemble  &  \xmark  & \xmark  \\
&  Ensemble-based &  $S$ & $S\times M$ & Centralized &  \cmark  &  \xmark \\
&  Model distillation &  $S$ &  $S$ & Ensemble  &  \cmark & \xmark  \\
\midrule[1pt]
\multicolumn{2}{c|}{\Ouralgo{}} &  $S$  & $S$ &  Centralized & \cmark & \cmark  \\
\bottomrule[1.5pt] 
\end{tabular}
\end{adjustbox}
\vspace{-0.1cm}
\label{tab:demystify-detail}
\end{table*}

\subsection{Data Heterogeneity in FL}\label{apdx:relatedNoniid}
The Non-IID data distribution is the notorious problem in FL, which severely harms the convergence rate and model performance of FL. The sharp Non-IID data makes local clients learn much different weights~\citep{karimireddy2019scaffold,woodworth2020minibatch}, resulting in heterogeneous feature~\citep{pmlr-v162-zhang22p,VHL} and classifiers~\citep{luo2021no}. Current methods that address Non-IID data problems include following typical directions.

Some works design new \textbf{FL optimizers} to stabilize and accelerate convergence~\citep{reddi2020adaptive,sun2023fedspeed}. \textbf{Personalized FL} aims to optimize different client models by learning knowledge from other clients and adapting to their own datasets~\citep{9743558,FedDF}. Distinguished from these works, how to alleviate data heterogeneity within extremely low communication costs is a new challenging problem, as clients have little possibility of communicating information with other clients. 

\textbf{Model regularization.} This direction proposes to add a penalty of distances between local trained models and the server model. FedProx~\citep{fedprox} directly uses the L2 distance between local models to the server model to constrain the local models not moving too far. SCAFFOLD~\citep{karimireddy2019scaffold} utilizes the historical local updates to correct update directions of local clients during local training. FedDyn~\citep{acar2021federated} dynamically updates the objective functions to ensure the local optima between devices are asymptotically consistent. FedIR~\citep{realgpd} claims that applying important weight to the client's local objectives helps to obtain an unbiased estimator of the global loss objective function.

\textbf{Feature calibration.} Some works focuses on align feature representations of different clients in similar spaces~\citep{li2021model,dong2022spherefed,VHL,10203389}. MOON~\citep{li2021model} adds the contrastive loss to between local and global models to learn a similar representation between clients, in which the global model acts as an intermediate agent to communicate between clients. It is found the local features of the same data largely shift between client models during local training~\citep{VHL}. To address this problem, virtual homogeneous learning~\citep{VHL} proposes to use a homogeneous dataset which can contain completely no information of original datasets, to calibrate the feature representations between clients. This technique improves the generalization performance and convergence speed of federated learning. SphereFed~\citep{dong2022spherefed} adds constraints on learned representations of input data to be in a unit hypersphere shared by clients. Besides, SphereFed discovers that the non-overlapped feature distributions for the same class lead to weaker consistency of the local learning targets from another perspective. The prototype~\citep{10203389} methods propose to utilize a pre-defined vector of each class in the representation space, then align client feature representations with such prototype, which is also called virtual feature transfer learning~\citep{VHL}. FedImpro~\citep{tang2024fedimpro} estimates and share the feature distribution to alleviate the gradient diversity and enhance high-level model training.

\textbf{Classifier calibration.} Due to the shifted features between clients, the classifier is usually trained with bias. And the final obtained classifier with FedAvg is normally prone to some specical classes. CCVR~\citep{luo2021no} is the first work to transmit the statistics of logits and label information of data samples to calibrate the classifier. In SphereFed~\citep{dong2022spherefed}, the classifier is fiexed with weights spanning the unit hypersphere, and calibrated by a mean squared loss. Some works also calibrate the classifier during or after training~\citep{dai2023tackling,Li_2023_ICCV,Kim_2023_ICCV}. 

\textbf{Optimization schemes.} 
From the optimization perspective, some methods regard local updates at clients as pseudo gradients~\citep{reddi2020adaptive} and design new FL optimizers to stabilize and accelerate convergence. FedNova~\citep{wang2020tackling} normalizes the local updates to reduce the inconsistency between the local and global optimization objective functions. FedAvgM~\citep{LDA} exploits the history updates of server model to avoid the overfits on the selected clients in each round. FedOpt~\citep{reddi2020adaptive} generalizes the centralized optimizers into FL scenario, and proposes FedAdaGrad, FedYogi, FedAdam. FedSpeed~\citep{sun2023fedspeed} utilizes a correction term on local updates to reduce the biases during training. FedSpeed also merges the stochastic gradient with a perturbation computed from an extra gradient ascent step in the neighborhood, to reduce the gradient heterogeneity.

\textbf{Data\&Feature sharing.}
The phenomenon of client drift primarily originates from data heterogeneity. To address this issue, researchers have discovered that sharing a subset of private data can markedly enhance convergence speed and generalization capability, as highlighted in~\citep{zhao2018federated}. However, this approach entails a compromise on the privacy of client data.

Consequently, to simultaneously mitigate data heterogeneity and uphold data privacy, a range of studies~\citep{5670948, NIPSAsimpleDPDataRelease, CompressiveLearninWithDP, johnson2018towards, li2024reflective, DataSynthesisDPMRF} have proposed the addition of noise to data. This method allows for the sharing of data while providing a certain level of privacy protection. Alternatively, other research efforts concentrate on disseminating synthetic data portions~\citep{jeong2018communication,long2021g,FLviaSyntheticData,hao2021towards} or data statistics~\citep{shin2020xor,yoon2021fedmix}, as opposed to raw data, in order to alleviate data heterogeneity.
FedDF~\citep{FedDF} employs external data and engages in knowledge distillation based on these data to facilitate the transfer of model knowledge between the server and clients. The fundamental concept of FedDF involves fine-tuning the aggregated model through knowledge distillation using newly shared data.

Additionally, to address feature shift, certain methodologies advocate for the sharing of features to enhance the convergence rate. Cronus~\citep{chang2019cronus} suggests the sharing of logits as a means to counteract poisoning attacks. CCVR~\citep{luo2021no} transmits the statistical data of logits to calibrate the final layer of federated models. CCVR~\citep{luo2021no} also shares parameters representative of local feature distribution. Importantly, this approach does not necessitate sharing the count of different labels with the server, thereby preserving the privacy of clients' label distribution. Furthermore, our method serves as a framework to leverage shared features in reducing gradient dissimilarity. The feature estimator employed need not be confined to the Gaussian distribution of local features; alternative estimators or even features from additional datasets, rather than private ones, may be utilized. In this direction, VHL~\citep{VHL} only requires to share the pure noise dataset, which can have completely no related information of the local private datasets, largely reducing the requirements of the shared datasets and can be exploited in practical scenarios.

\textbf{Personalized FL.} 
Personalized Federated Learning (PFL) aims to enable clients to optimize distinct personal models that can absorb knowledge from other clients and tailor it to their individual datasets, as detailed in~\citep{9743558}. The process of knowledge transfer in personalization primarily involves the introduction of personalized parameters~\citep{liang2020think,thapa2020splitfed,9708944}, or the employment of knowledge distillation~\citep{FedGKT2020, FedDF, li2019fedmd,bistritz2020distributed} utilizing shared local features or supplementary datasets.

However, due to the tendency of personalized federated models to prioritize optimizing local objective functions, they often do not achieve generic performance (as evaluated on a global test dataset) that is on par with conventional Federated Learning (FL) methodologies~\citep{chen2021bridging}. Given our primary objective of learning an enhanced generic model, we have chosen not to include comparisons and improve performance of personalized FL algorithms in our work. While achieving extremely low communication costs in PFL can be an interesting future work.

\subsection{One-shot FL}\label{apdx:relatedOneshot}
Current FL methods with communication costs of only one model size are referred as one-shot FL, different one-shot learning, the ``one-shot'' here means one communicating round~\citep{zhang2022dense,guha2019one,li2020practical,zhou2020distilled,dennis2021heterogeneity}. Thus, the communication cost is limited as the model size $\Mcal$ for each client, less than FedAvg-style algorithms for $R$ times. While there are works that introduce additional communication costs to improve the performance of one-shot FL. Existing one-shot FL works can be categorized into average-based~\citep{jhunjhunwala2023towards}, ensemble-based~\citep{wang2023datafree} or model distillation based~\citep{heinbaugh2023datafree,diao2023towards}, 

\textbf{Average-based} methods include the basic baseline, one-shot FedAvg, which shows a severely bad performance, and some other advanced average methods~\citep{matena2022merging,jin2023dataless,yadav2023tiesmerging}. FedFisher~\citep{jhunjhunwala2023towards} proposes to utilize fisher information to avoid harming knowledge of local models when averaging on the server. Note that the similar methodology of using fisher information is also utilized to enhance the FedAvg~\citep{qu2022generalized} or bayesian FL~\citep{liu2023bayesian,al2020federated}, which belongs to multi-round FL. All of them aims to avoid forgetting local learned knowledge like the classical exploitation of fisher information in continual learning~\citep{PNASforget}. 
Matching permutations between the weights of different models~\citep{ainsworth2022git, wang2020Federated} is another advanced method for model averaging.
Linear mode connectivity~\citep{pmlr-v119-frankle20a,neyshabur2020being,ainsworth2022git} helps to explain to the part of the success of model averaging. However, due to the highly non-linear structure of deep neural networks, it is extremely difficult to find a method to directly average local clients through one round to obtain a perfect server model. Therefore, one-shot average method usually fails to achieve a good performance.

\textbf{Ensemble-based} methods are based on ensemble learning~\citep{guha2019one}. Intuitively, the server aggregates multiple local client models. Then, the direct way of deploying these models is to average outputed logits of them and make predictions~\citep{wang2023datafree}. Some methods propose to find a better model selection method to select local models that might be more familiar with the given inputs~\citep{wang2023datafree}. Some methods utilize prototype data~\citep{diao2023towards} or generated datasets~\citep{heinbaugh2023datafree} to conduct better model selection in ensemble learning.

\textbf{Model distillation} methods are built upon the ensemble models~\citep{guha2019one,li2020practical}. The core motivation is that the ensemble models occupy too much storage and require much extra computation costs. Thus~\citep{li2020practical} involved a public dataset for training. As a replacement of using global data,~\citep{zhou2020distilled} transmits the distilled local datasets to server for model distillation.~\citep{dennis2021heterogeneity} clusters clients and communicates the mean data of each cluster. These methods require extra datasets, which may be impractical in some data-sensitive scenarios like medical and education data. And the large storage costs still exist when many clients upload their models for ensemble learning.

Note that there is also some work~\citep{park2021few} that proposes to find a good initial model of FL in few communication rounds. Thus the convergence rate of FedAvg can be accelerated by this good initialization~\citep{nguyen2022begin,he2021fedcv}. The objective of this work is different from one-shot FL or ours.

\subsection{Communication compressed FL}\label{apdx:compressFL}
Different from one-shot FL which reduces the communication frequency, communication compression methods aim to reduce the communication size in each round. Typical communication-compression methods~\citep{tang2020survey} include sparsification~\citep{bibikar2022federated,jiang2022model,tang2020communication,GossipFL}, quantization~\citep{reisizadeh2020fedpaq,gupta2023quantization} and low-rank decomposition~\citep{nguyen2024towards, kone2016}.

\textbf{Sparsification.} Studies in~\citep{DoCoFL,bibikar2022federated, qiu2022zerofl, GossipFL} have introduced a significant level of sparsity in the local model training stage, effectively reducing the number of parameters that need to be transmitted. The works represented in~\citep{frankle2018lottery} have put forth the Lottery Ticket Hypothesis, suggesting the existence of trainable sub-networks within over-parameterized networks that can be independently trained without accuracy loss. Inspired by this concept, LotteryFL~\citep{9708944} and FedLTN~\citep{mugunthan2022fedltn} aim to identify and exchange these personalized lottery ticket networks between the server and clients. Moreover, Hetero-FLASH~\citep{babakniya2022federated} employs adaptive sparsity, with objectives extending beyond identifying the optimal sub-network, to fully leveraging clients’ resources. In contrast to focusing on personalized Federated Learning (FL), this paper primarily considers generic FL, where all clients share the same model structure.

\textbf{Quantization.} Orthogonally to Sparsification, quantization emerges as an additional pivotal strategy to alleviate the communication bottleneck in Federated Learning. This method represents model updates with fewer bits than the conventional 32 or 64 bits, thus reducing numerical precision. FedPAQ~\citep{reisizadeh2020fedpaq} adopts periodic averaging of low-bit representations of local model updates to minimize both the frequency of communication and the overhead per round. Advancements in this realm have been furthered by~\citep{gupta2023quantization}, which introduces a variant of Quantization-Aware Training (QAT) robust to multiple bit-widths, eliminating the necessity for retraining in the FL context.

\textbf{Low-rank decomposition.} Concurrently, research in~\citep{nguyen2024towards, kone2016} has utilized a low-rank decomposition of matrices to cultivate sparse models. Specifically, the weights of local trained models are decomposed into smaller matrices. Then these smaller matrices are communicated to the server for recovering the original weights.

While these methods also reduce the communication costs, they are far from reducing costs at an extremely low level of the one-shot FL and ours. Because these methods still require many even more communication rounds than FedAvg to achieve training convergence.



\subsection{Mutual Information}\label{apdx:relatedMI}
The mutual information (MI) has garnered increasing attention in recent years with its explanation of how deep neural networks learn intermediate representations of the raw data. The Information Bottleneck (IB) principle~\citep{shwartz2017opening,saxe2019information,tishby99information} provides insights into the training dynamics of deep neural networks. Specifically, the neural network reduces the mutual information between raw data and representations layer by layer, while maximizing the mutual information between labels and representations.~\citep{achille2018emergence} introduces the nuisance variable into the mutual information, and proposes that mutual information between representation and the nuisance should be as less as possible. And it is proved that the empirical risk minimization (ERM) with stochastic gradient descent (SGD) has implicitly achieved the IB principle~\citep{achille2018emergence}. Some methods in unsupervised learning exploit maximization of mutual information~\citep{oord2018representation, tian2020makes, hjelm2018learning} to enhance the feature representation. The contrastive learning~\citep{chen2020simple,he2020momentum} maximizes the mutual information across varying views of the same input.



\subsection{Out-of-Distribution Generalization}\label{apdx:OOD-gene}
Out-of-Distribution (OOD) generalization refers to the model performance on the unseen data distribution, which is called OOD data~\citep{liu2021towards,chen2023understanding,ahuja2021invariance}. The spurious features widely exist in the real-world datasets, like the textures, shapes, colors of objects~\citep{hendrycks2021natural,geirhos2018imagenet,geirhos2020shortcut,wangex,luo2023ai}. Classification with original labels with these properties often leads overfitting on these variables instead of learning the real mapping relationship $X \to Y$. It is found that the out-of-distribution generalization performance of a model learned by ERM~\citep{chen2023understanding,ye2021towards} is connected with the spurious features. It is found that the over-parameterization, increasing model size well beyond the point of zero training error, may hurt the test error on some data samples, while it improving the average test error on all data samples~\citep{sagawa2020investigation}. These aggravated test errors are called worst-group error~\citep{liu2021towards}.

Invariant risk minimization (IRM)~\citep{arjovsky2019invariant,ahuja2021invariance} is proposed to address inheriting spurious correlations in trained models. It is shown that exploiting invariant causal relationships between datasets gathered from multiple environments rather than relying on varying spurious relationships appearing in isolated local datasets, help to improve the robustness of the learned predictors. ~\citep{zhang2021can} proposes that there exists a sub-network hidden in the full neural network that is unbiased functional (not focusing on spurious correlation), thus achieving better OOD performance.

However, some studies show that there is no clear winner between ERM and IRM when covariate shift happens~\citep{ahuja2020empirical}, and ERM is still the state-of-the-art on many problems of OOD generalization~\citep{gulrajani2020search}.

The colored MNIST (CMNIST) is also called anti-causal CMNIST proposed by~\citep{arjovsky2019invariant,ahuja2020empirical} is often used to study the OOD performance of different training methods. In our work, we directly add different geometry shapes with random colors on local clients as the spurious features (also called adversarial samples~\citep{kaiwen,margin5,ho2020contrastive}) to explore how isolated local training and our algorithm are influenced by them.

It is proven that the self-supervised or unsupervised training with contrastive learning~\citep{ho2020contrastive,naseer2020self} can help reduce the overfitting on spurious features. And dropout~\citep{Dropout} also helps learning spurious relationships between label and the spurious features~\citep{vivek2020single}.

\section{Details of Experiment Configuration}\label{apdx:DetailedExp}

\subsection{Hyper Parameters}
\begin{table*}[h]
\centering
\caption{Learning rate of all experiments.} 
\vspace{1pt}
\begin{tabular}{c|cccccc}
\toprule[1.5pt]
Dataset & MNIST & FMNIST & CIFAR10 & SVHN  & CIFAR100 & Tiny-Imagenet \\
\midrule[1pt]
$a=0.1$ & 0.001 & 0.001  & 0.01    & 0.001 & 0.01 & 0.01   \\
$a=0.3$ & 0.001 & 0.01   & 0.01    & 0.01  & 0.01 & 0.01   \\
$a=0.5$ & 0.001 & 0.01   & 0.01    & 0.01  & 0.01 & 0.01   \\
\bottomrule[1.5pt] 
\end{tabular}
\label{tab:LearningRate}
\end{table*}

The learning rate configuration has been listed in Table~\ref{tab:LearningRate}. We report the best results and their learning rates (searched in $\left \{0.0001, 0.001, 0.01, 0.1 \right \}$). During local updating, the optimizer is SGD with 0.9 momentum following most FL baselines~\citep{}


\subsection{Hardware and software}\label{apdx:exp-hardware-software}
The FedAvg baseline is conducted based on the standard commonly used FL library FedML~\citep{chaoyanghe2020fedml,tang2023fedml}. While other OFL baselines are implemented following~\citep{zhang2022dense}. All experiments are conducted based on NVIDIA 2080 Ti or NVIDIA V100 GPU for several hours. Users just need one single GPU to run these experiments.

\subsection{Mutual Information Estimation}\label{apdx:EstMI}
Due to the extremely high dimension of intermediate features, MI estimation is very difficult~\citep{belghazi2018mutual}. Simple neural networks may fail to accurately estimate MI. Thus, we follow ~\citep{wang2020revisiting}, to use the reconstruction error from $\bm{h}$ to $\bm{x}$ to estimate the $I(\bm{h}, \bm{x})$, and the classification error to estimate $I(\bm{h}, y)$. The details are provided as follows.

\textbf{Estimating $I(\bm{h}, \bm{x})$.}
Assume that $\mathcal{R}(\bm{x}|\bm{h})$ denotes the expected error for reconstructing $\bm{x}$ from $\bm{h}$. It has been widely known that $\mathcal{R}(\bm{x}|\bm{h})$ follows $I(\bm{h}, \bm{x})=H(\bm{x})-H(\bm{x}|\bm{h})\geq H(\bm{x})-\mathcal{R}(\bm{x}|\bm{h})$, where $H(\bm{x})$ denotes the marginal entropy of $\bm{x}$, as a constant~\citep{hjelm2018learning}. We estimate $I(\bm{h}, \bm{x})$ by training a decoder parameterized by $w$ to reconstruct the original image $x$, namely $I(\bm{h}, \bm{x})\approx\mathop{\max}_{w}[H(x)-\mathcal{R}_{w}(x|h)]$.  For estimating $I(\bm{h}, \bm{x})$, the decoders are multiple up-sampling convolutional layers following~\citep{wang2020revisiting}.


\textbf{Estimating $I(\bm{h}, y)$.}
Since $I(h, y)=H(y)- H(y|h)=H(y)-\mathbb{E}{(h, y)}[-\text{log}\ p(y|h)]$, we can directly train an auxiliary classifier $q{\psi}(y|h)$ with parameters $\psi$ to approximate $p(y|h)$, such that we have $I(h, y) \approx \mathop{\max}{\psi} {H(y)-\mathbb{E}{h}[\sum_{y}-p(y|h)\text{log}\ q_{ \psi}(y|h)]}$. To summarize, given the split features $H^k$ at layer $k$, we freeze previous layers that with index $i \leq k$, and train a new inserted linear layer as classifier as to estimate $I(h, y)$. For estimating $I(h, x)$, the decoder follows ~\citep{wang2020revisiting}.



\subsection{Linear Separability}\label{apdx:EstLP}
Following~\citep{alain2016understanding,NIPS2017_10ce03a1}, for each layer $k$ to be examined, we stitch and train a linear classifier $MLP_k$ following it, while freezing previous layers. Thus, the linear separability of feature $H^k$ is shown by the classification error of $MLP_k$. The MLP is trained by 10 epochs.

\subsection{Backdoored Datasets}\label{apdx:backdoor}
\begin{figure*}[htb!] 
    \setlength{\abovedisplayskip}{-2pt}
    \subfigbottomskip=-1pt
   \centering
    \subfigure{\includegraphics[width=0.49\linewidth]{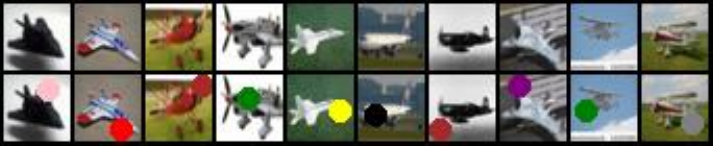}}
    \subfigure{\includegraphics[width=0.49\linewidth]{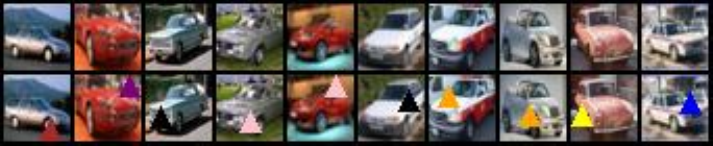}}
    \subfigure{\includegraphics[width=0.49\linewidth]{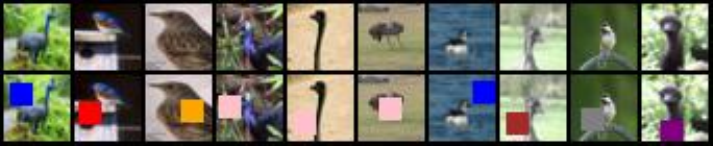}}    \subfigure{\includegraphics[width=0.49\linewidth]{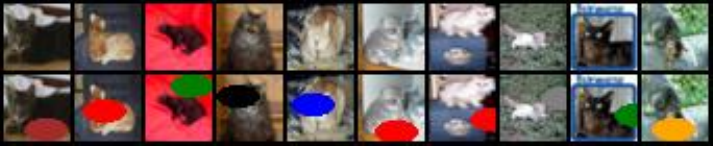}}
    \subfigure{\includegraphics[width=0.49\linewidth]{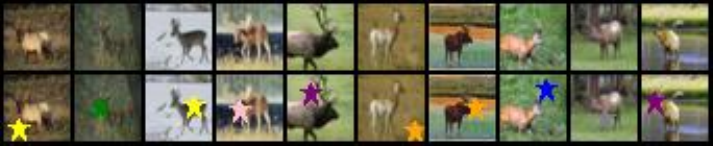}}
    \subfigure{\includegraphics[width=0.49\linewidth]{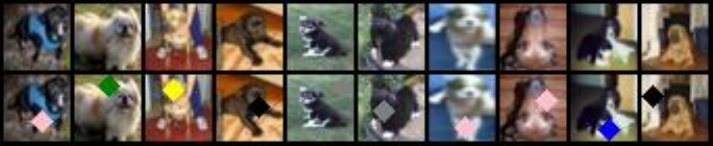}}
    \subfigure{\includegraphics[width=0.49\linewidth]{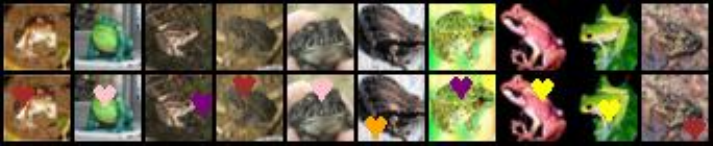}}
    \subfigure{\includegraphics[width=0.49\linewidth]{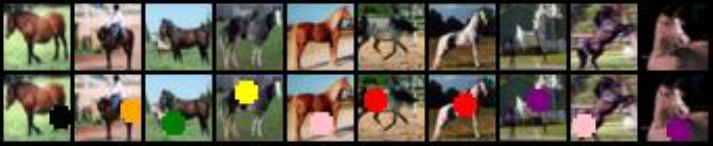}}
    \subfigure{\includegraphics[width=0.49\linewidth]{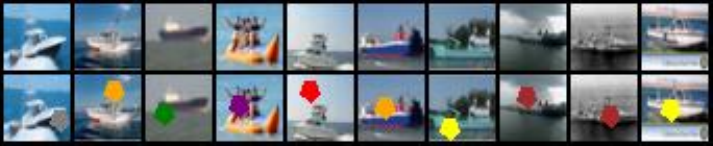}}
    \subfigure{\includegraphics[width=0.49\linewidth]{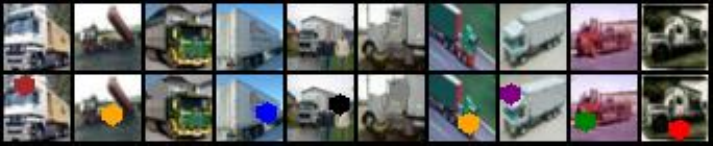}}
    \caption{Each row is a class of original (upper) and backdoored (lower) images of CIFAR-10. The shapes are added on images according to label indexes.} 
    \label{fig:BackdoorImages}
\vspace{-0.0cm}
\end{figure*}

Figure~\ref{fig:BackdoorImages} shows the original images and the backdoored images of CIFAR-10. The shapes are added on images according to label indexes but with random colors. By training on backdoored images, the local model is easily to fit on the shapes instead of original images. Each shape occupies $10\times10$ image size. Table~\ref{tab:backdooracc}  provides the test accuracy of different methods training with backdoored CIFAR-10, showing that the performance is severely harmed by the backdoored datasets. While \Ouralgo{} is not severely affected by the backdoor.

\section{More Results}\label{apdx:moreresults}
To validate the effect of \Ouralgo{} when training heterogeneous models with more clients, we further conduct experiments with $M=10$ and compare results with $M=5$ as shown in Table~\ref{tab:modelHetero-difNclients}. Results show that the \Ouralgo{} well supports training heterogeneous models, of which the performance is still comparable with Ensemble FL, but requires much less storage cost than Ensemble.

\subsection{Heterogeneous Model with Different Number of Clients}
Table~\ref{tab:modelHetero-difNclients} shows how number of clients influences the performance of \Ouralgo{} with average or \conv{} as adapter. Results show that with increased $M=10$, \Ouralgo{} still provide benefits to training heterogeneous model.

\begin{table}[htb!]
\centering
\caption{Accuracy with \Ouralgo{} with \conv{} or averaging to support heterogeneous model design on CIFAR-10 with different number of clients. }
\begin{tabular}{c|cc}
\toprule
\# Clients & $M=5$       & $M=10$      \\
\midrule
\rowcolor{greyL} Ensemble & 79.91 & 77.25  \\
\midrule
\Ouralgo{}              & 81.15 & \textbf{78.28}  \\
\Ouralgo{} (Avg)        & 80.35 & 76.52  \\
\Ouralgo{} Hetero       & \textbf{82.71} & 76.47  \\
\Ouralgo{} (Avg) Hetero & 79.74 & 75.90  \\
\bottomrule
\end{tabular}
\label{tab:modelHetero-difNclients}
\end{table}

\subsection{Comparisons with FedMA}\label{apdx:compare-FedMA}
FedMA~\citep{wang2020Federated} does not support training ResNet-18 and aggregating multi-layers blocks. Thus, we compare it with FedAvg and our methods on training VGG-9 on CIFAR-10. For FedAvg, we run it for 10 communication rounds. For ensemble FL methods, the training epoch is set as 200 and communication with only one round. For fair comparison, to keep the same computation burden, we divide the 200 epochs to each communication round in \Ouralgo{} and FedMA. Thus, FedMA, \Ouralgo{} and Ensemble FL have the same communication costs and $10\times$ less than FedAvg. Note that the \Ouralgo{} can have different $K$ to decide its communication rounds, while FedMA has a fixed number of communication rounds, i.e. the number of layers in VGG-9. Thus, the total computation and communication costs of different baselines are same except for FedAvg. Results in Table~\ref{tab:compareFedMA} shows that \Ouralgo{} successfully outperforms other methods. Note that to achieve the comparable performance to hundreds-of-rounds FedAvg, the FedMA actually requires communicating multiple model size. UNder the limited communication constraints, the matching and averaging ways in FedMA show inferior performance than the feature concatenation and merging as used in \Ouralgo{}.

\begin{table}[htb!]
\centering
\caption{Comparing accuracy on CIFAR-10 with FedMA. }
\begin{tabular}{c|ccc}
\toprule
Algorithm   & $a=0.1$  & $a=0.3$  & $a=0.5$  \\
\midrule
FedAvg &  21.19  &  26.16 &  38.63 \\
FedMA  &  66.28 &  71.95 &  73.34   \\
Ensemble &  65.72 &  67.29  &  69.58    \\
\Ouralgo{} $K=2$ & 73.99  &  75.58  &  76.66 \\
\Ouralgo{} $K=3$ & 72.31  &  75.29  &  76.10 \\
\Ouralgo{} $K=4$ & 71.34  &  76.87  &  75.71  \\
\bottomrule
\end{tabular}
\label{tab:compareFedMA}
\end{table}

\subsection{Different Feature Fusion Methods}\label{apdx:featurefusion}

\begin{table}[htb!]
\centering
\caption{Accuracy with \Ouralgo{} with \conv{} or averaging as adapters of different non-IID degree on CIFAR-10. }
\begin{adjustbox}{max width=\linewidth}
\begin{tabular}{c|ccc}
\toprule
non-IID degree & $a=0.1$        & $a=0.3$        & $a=0.5$        \\
\midrule
\rowcolor{greyL} Ensemble               & 57.5           & 77.35          & 79.91          \\
\midrule
\Ouralgo{} $K=2$       & 70.85          & 81.41          & \textbf{84.34} \\
\Ouralgo{} $K=4$       & \textbf{73.79} & \textbf{84.58} & 81.15          \\
\Ouralgo{} $K=8$       & 70.46          & 80.7           & 74.99          \\
\Ouralgo{} (Avg) $K=2$ & 70.79          & 81.5           & 83.56          \\
\Ouralgo{} (Avg) $K=4$ & 68.08          & 71.49          & 80.35          \\
\Ouralgo{} (Avg) $K=8$ & 71.58          & 81.87          & 83.29       \\ \bottomrule
\end{tabular}
\end{adjustbox}
\label{tab:conv1x1VSavg}
\end{table}

We verify the effect of using \conv{} or simply averaging features as feature fusion in Table~\ref{tab:conv1x1VSavg}. Results shows that using \conv{} is generally better than averaging. With the decreased non-IID degree, the gap between \conv{} and averaging is smaller, demonstrating that more similar features require less feature adaptation through learning a mapping i.e. training a \conv{}.

\subsection{Communication Cost Comparison}\label{apdx:commcost}

\begin{table}[htb!]
\centering
\caption{Communication costs. For different number of clients, the number of basic channels in ResNet-18 of \Ouralgo{} is set as 32, 20, 14, 9 with $M\in\cbr{5, 10, 20, 50}$, respectively.}
\begin{adjustbox}{max width=\linewidth}
\begin{tabular}{c|cccc}
\toprule
\# Clients        & $M=5$    & $M=10$   & $M=20$   & $M=50$    \\
\midrule
\rowcolor{greyL} Single Model     & 42.66MB  & 42.66MB  & 42.66MB  & 42.66MB   \\
\midrule
FedAvg (10 rounds) & 2133.1MB & 4266.2MB & 8532.4MB & 21331.0MB \\
Other OFLs & 213.31MB & 426.62MB & 853.24MB & 2133.10MB \\
Ensemble         & 213.31MB & 426.62MB & 853.24MB & 2133.10MB \\
\Ouralgo{} $K=2$ & 268.55 MB & 423.2 MB & 842.6 MB & 2274.0 MB   \\
\Ouralgo{} $K=4$ & 276.9 MB & 449.2 MB & 944.4 MB & 2931.5 MB   \\
\Ouralgo{} $K=8$ & 340.4 MB & 647.7 MB & 1722.2 MB & 7954.0 MB  \\
\Ouralgo{} (Avg) & \textbf{266.6 MB} & \textbf{416.6 MB} & \textbf{816.6 MB} & \textbf{2109.0 MB} \\
\bottomrule
\end{tabular}
\end{adjustbox}
\label{tab:commcost}
\vspace{-0.1cm}
\end{table}

We provide the detailed communication costs of different methods in Table~\ref{tab:commcost}. The Other OFLs refer to advanced OFL method including DENSE~\citep{zhang2022dense}, data-free KD methods DAFL~\citep{chen2019data} and ADI~\citep{yin2020dreaming}. Table~\ref{tab:commcost} shows that \Ouralgo{} does not increase the communication costs, while largely improving the model performance.

\subsection{Higher Heterogeneity and More Baselines}\label{apdx:higherHetero}
Table~\ref{tab:highHeteo} provides more results of the higher heterogeneity ($a=0.05$) and more baselines including DENSE~\citep{zhang2022dense} and CoBoosting~\citep{dai2024enhancing}. Results show that the CoBoosting can improve the performance than other baseline methods but fail to outperform \Ouralgo{}.

Both CoBoosting~\citep{dai2024enhancing} and FEDCVAE-KD~\citep{heinbaugh2023datafree} focus on exploiting knowledge distillation methods to improve the performance of global models, while our method focuses on how to aggregate models together. Thus, the knowledge distillation is orthogonal with our method and may be utilized to enhance \Ouralgo{}. For example, one can consider running \Ouralgo{} first to obtain a fused global model. Then, this model can be used to conduct knowledge distillation to guide local model training with the \Ouralgo{} once again.

\begin{table}[htb!]
\centering
\caption{Results of higher data heterogeneiry ($a=0.05$).}
\begin{adjustbox}{max width=\linewidth}
\begin{tabular}{c|ccccc}
\toprule
Datasets & MNIST & FMNIST & SVHN & CIFAR-10 & CIFAR-100 \\
\midrule
\rowcolor{greyL} Ensemble  & 58.06  & 66.19  & 62.22  & 53.33   & 32.25  \\
\midrule
FedAvg    & 46.35  & 20.07  & 39.41  & 17.49   & 6.45   \\
FedDF     & 80.73  & 44.73  & 60.79  & 37.53   & 16.07  \\
F-ADI     & 80.12  & 42.25  & 56.58  & 36.94   & 13.75  \\
F-DAFL    & 78.49  & 41.66  & 59.38  & 37.82   & 15.79  \\
DENSE     & 81.06  & 44.77  & 60.24  & 38.37   & 16.17  \\
CoBoosting& 93.93  & 50.62  & 65.40  & 47.20   & 19.24  \\
\Ouralgo{} $K=2$ & 95.23  & 83.23  & 75.08  & 46.38   & 29.98  \\
\Ouralgo{} $K=4$ & \textbf{95.37}  & \textbf{83.65}  & \textbf{75.53}  & \textbf{51.59}   & \textbf{32.71} \\
\bottomrule
\end{tabular}
\end{adjustbox}
\label{tab:highHeteo}
\vspace{-0.1cm}
\end{table}

\section{Limitation}\label{apdx:lim}
\textbf{Finding invariant features.} With our analysis in Section~\ref{sec:analysis}, it will be useful to choose the invariant features $\INV{m}$ during local training for client $n$. Currently, under the communication and privacy constrains, it is further difficult to identify which features are spurious. We left this as the future work to explore. We do not explore and exploit this technique in this paper, as it is orthogonal to our core innovation technique, augmenting features by layer-wise model fusion. Nevertheless, by manually crafting spurious correlations by adding backdoored data samples into local dataset, we empirically prove \Ouralgo{} can effectively avoid the influences of spurious features. 


\textbf{Limited feature adaptation.} In this work, as an initial trial, we only explore using averaging and \conv{} as the feature adaptation. Further methods can consider exploring better feature adaptation methods like using non-linear models.

\textbf{Security issues.} In this work, we do not explicitly consider the security issue. However, the vulnerability to attacks of \Ouralgo{} will not be higher than previous multi-round FedAvg, which requires many more communication rounds to achieve the same model performance as \Ouralgo{}. For example, FedAvg may require more than 100 rounds to achieve 70\% test accuracy as \Ouralgo{} with $2\sim 4$ rounds, which introduces more communicated information and a higher possibility of attacks. And \Ouralgo{} has the same communication size with other OFL methods.

However, while the size of the communication does not increase, \Ouralgo{} increases the number of communication rounds compared to other OFL methods. There are some possible mitigations to address the security issues:
\begin{itemize}[leftmargin=*]
    \item \textbf{Adversarial attacks.} Some malicious clients might upload adversarial modules or backdoored modules that are used to misguide the aggregated model to generate incorrect or handcrafted predictions. For these attacks, the possible solution is to detect and reject such malicious uploading also through the lens of causality. Specifically, some images with the invariant features can be fed into the uploaded modules to see whether the output feature can be used to correctly classify images;
    \item \textbf{Model inversion or Membership attack.} Some malicious clients or the server may consider to conduct model inversion or membership attack to obtain the raw data of clients, thus threatening the user privacy. In this case, the learned module can be protected with differential privacy to enhance its security.
\end{itemize}

\textbf{Multiple communication rounds.} To enable concatenating local modules together and training based on aggregated features from these modules, multiple communication rounds are required in \Ouralgo{}. In this sense, \Ouralgo{} belongs to the few-shot FL. However, the communication cost of \Ouralgo{} is as same as other OFL methods, which is the main claim in the introduction (extremely low communication costs).

\newpage
\section*{NeurIPS Paper Checklist}

\begin{enumerate}

\item {\bf Claims}
    \item[] Question: Do the main claims made in the abstract and introduction accurately reflect the paper's contributions and scope?
    \item[] Answer: \answerYes{} 
    \item[] Justification: The main claims made in the abstract and introduction accurately reflect the paper's contributions and scope.

\item {\bf Limitations}
    \item[] Question: Does the paper discuss the limitations of the work performed by the authors?
    \item[] Answer: \answerYes{} 
    \item[] Justification: The limitations have been discussed in Section~\ref{apdx:lim}. The communication cost and storage costs are compared in Table~\ref{tab:memory} and ~\ref{tab:commcost}. The scalability on datasets and clients, computation efficiency, and whether supporting personalized model design have been discussed in Experiment Section~\ref{sec:exp}.

\item {\bf Theory Assumptions and Proofs}
    \item[] Question: For each theoretical result, does the paper provide the full set of assumptions and a complete (and correct) proof?
    \item[] Answer: \answerNA{} 
    \item[] Justification: We build the causal graph without giving the theorem. The proof of Lemma~\ref{lemma:invariance} comes from ~\citep{achille2018emergence}

    \item {\bf Experimental Result Reproducibility}
    \item[] Question: Does the paper fully disclose all the information needed to reproduce the main experimental results of the paper to the extent that it affects the main claims and/or conclusions of the paper (regardless of whether the code and data are provided or not)?
    \item[] Answer: \answerYes{} 
    \item[] Justification: We have provided the experiment setting, specific baselines (Section~\ref{sec:exp}), hyper-parameters (Section~\ref{apdx:DetailedExp}), the algorithm details (Algorithm~\ref{apdx:algo}, hardware and software details (Section~\ref{apdx:exp-hardware-software}). We have provided our open-sourced code link.

\item {\bf Open access to data and code}
    \item[] Question: Does the paper provide open access to the data and code, with sufficient instructions to faithfully reproduce the main experimental results, as described in supplemental material?
    \item[] Answer: \answerYes{} 
    \item[] Justification: We have provided our open-sourced code link. The library has provided the experiment and code instructions. And the datasets are public datasets as listed in Section~\ref{sec:exp}, which can be downloaded online, or using Pytorch.

\item {\bf Experimental Setting/Details}
    \item[] Question: Does the paper specify all the training and test details (e.g., data splits, hyperparameters, how they were chosen, type of optimizer, etc.) necessary to understand the results?
    \item[] Answer: \answerYes{} 
    \item[] Justification: We have provided the experiment setting, specific baselines, optimizers(Section~\ref{sec:exp}), hyper-parameters (Section~\ref{apdx:DetailedExp}), the algorithm details (Algorithm~\ref{apdx:algo}, hardware and software details (Section~\ref{apdx:exp-hardware-software}).

\item {\bf Experiment Statistical Significance}
    \item[] Question: Does the paper report error bars suitably and correctly defined or other appropriate information about the statistical significance of the experiments?
    \item[] Answer: \answerNo{} 
    \item[] Justification: We do not report the error bars in our experiments.

\item {\bf Experiments Compute Resources}
    \item[] Question: For each experiment, does the paper provide sufficient information on the computer resources (type of compute workers, memory, time of execution) needed to reproduce the experiments?
    \item[] Answer: \answerYes{} 
    \item[] Justification: We have provided the experiment setting (Section~\ref{sec:exp}) and hardware and software details (Section~\ref{apdx:exp-hardware-software}). 
    
\item {\bf Code Of Ethics}
    \item[] Question: Does the research conducted in the paper conform, in every respect, with the NeurIPS Code of Ethics \url{https://neurips.cc/public/EthicsGuidelines}?
    \item[] Answer: \answerYes{} 
    \item[] Justification: This work does not incorporate any ethic concerns of NeurIPS. The datasets and models are commonly used in the community, and the method does not incorporate potential concerns.

\item {\bf Broader Impacts}
    \item[] Question: Does the paper discuss both potential positive societal impacts and negative societal impacts of the work performed?
    \item[] Answer: \answerYes{} 
    \item[] Justification: We have discussed the broader impact in Section~\ref{apdx:impact}.
    
\item {\bf Safeguards}
    \item[] Question: Does the paper describe safeguards that have been put in place for responsible release of data or models that have a high risk for misuse (e.g., pretrained language models, image generators, or scraped datasets)?
    \item[] Answer: \answerNA{} 
    \item[] Justification: This paper poses no such risks.

\item {\bf Licenses for existing assets}
    \item[] Question: Are the creators or original owners of assets (e.g., code, data, models), used in the paper, properly credited and are the license and terms of use explicitly mentioned and properly respected?
    \item[] Answer: \answerYes{} 
    \item[] Justification: The datasets and baselines, used libraries are properly credited.

\item {\bf New Assets}
    \item[] Question: Are new assets introduced in the paper well documented and is the documentation provided alongside the assets?
    \item[] Answer: \answerYes{} 
    \item[] Justification: The datasets and baselines, used libraries are well documented and cited.

\item {\bf Crowdsourcing and Research with Human Subjects}
    \item[] Question: For crowdsourcing experiments and research with human subjects, does the paper include the full text of instructions given to participants and screenshots, if applicable, as well as details about compensation (if any)? 
    \item[] Answer: \answerNA{} 
    \item[] Justification: The paper does not involve crowdsourcing nor research with human subjects.

\item {\bf Institutional Review Board (IRB) Approvals or Equivalent for Research with Human Subjects}
    \item[] Question: Does the paper describe potential risks incurred by study participants, whether such risks were disclosed to the subjects, and whether Institutional Review Board (IRB) approvals (or an equivalent approval/review based on the requirements of your country or institution) were obtained?
    \item[] Answer: \answerNA{} 
    \item[] Justification: The paper does not involve crowdsourcing nor research with human subjects.

\end{enumerate}

\end{document}